\documentclass[10pt,twocolumn,letterpaper]{article}

\usepackage[pagenumbers]{cvpr} 
\usepackage{multirow}
\usepackage{times}
\usepackage{epsfig}
\usepackage{amsmath}
\usepackage{amssymb}
\usepackage[hyphens]{url}
\usepackage{tabularx}
\usepackage{pifont}
\usepackage[flushleft]{threeparttable}
\usepackage[accsupp]{axessibility}  
\graphicspath{{../figures/}}
\usepackage{xcolor}

\newcommand{\jvg}[1]{[{\bf \color{orange} JvG: #1}]}
\newcommand{\yclin}[1]{[{\color{blue} yclin: #1}]}

\newcommand{\Eq}[1]{Eq.~(\ref{eq:#1})}
\newcommand{\eq}[1]{\Eq{#1}}
\newcommand{\fig}[1]{Fig.~\ref{fig:#1}}
\newcommand{\tab}[1]{Tab.~\ref{tab:#1}}

\usepackage{graphicx}
\usepackage{amsmath}
\usepackage{amssymb}
\usepackage{booktabs}
\usepackage{comment}

\newcommand{\norm}[1]{\left\lVert#1\right\rVert}

\usepackage[pagebackref,breaklinks,colorlinks]{hyperref}

\usepackage[capitalize]{cleveref}
\crefname{section}{Sec.}{Secs.}
\Crefname{section}{Section}{Sections}
\Crefname{table}{Table}{Tables}
\crefname{table}{Tab.}{Tabs.}

\def\cvprPaperID{4266} 
\def\confName{CVPR}
\def\confYear{2022}

\begin{document}

\title{Deep vanishing point detection: Geometric priors make dataset variations vanish}

\author{Yancong Lin, Ruben Wiersma, Silvia L. Pintea, Klaus Hildebrandt,
Elmar Eisemann, and Jan C. van Gemert\\
Delft University of Technology, The Netherlands\\
}




\maketitle

\begin{abstract}
Deep learning has improved vanishing point detection in images.
Yet, deep networks require expensive annotated datasets trained on costly hardware and do not generalize to even slightly different domains, and minor problem variants. 
Here, we address these issues by injecting deep vanishing point detection networks with prior knowledge. 
This prior knowledge no longer needs to be learned from data, saving valuable annotation efforts and compute, unlocking realistic few-sample scenarios, and reducing the impact of domain changes.
Moreover, the interpretability of the priors allows to adapt deep networks to minor problem variations such as switching between Manhattan and non-Manhattan worlds. 
We seamlessly incorporate two geometric priors: (i) Hough Transform -- mapping image pixels to straight lines, and (ii) Gaussian sphere -- mapping lines to great circles whose intersections denote vanishing points. 
Experimentally, we ablate our choices and show comparable accuracy to existing models in the large-data setting. We validate our model\rq s improved data efficiency, robustness to domain changes, adaptability to non-Manhattan settings.
\end{abstract}
\section{Introduction}

\begin{figure}[t!]
    \centering
    \begin{tabular}{c}
    \includegraphics[width=0.45\textwidth]{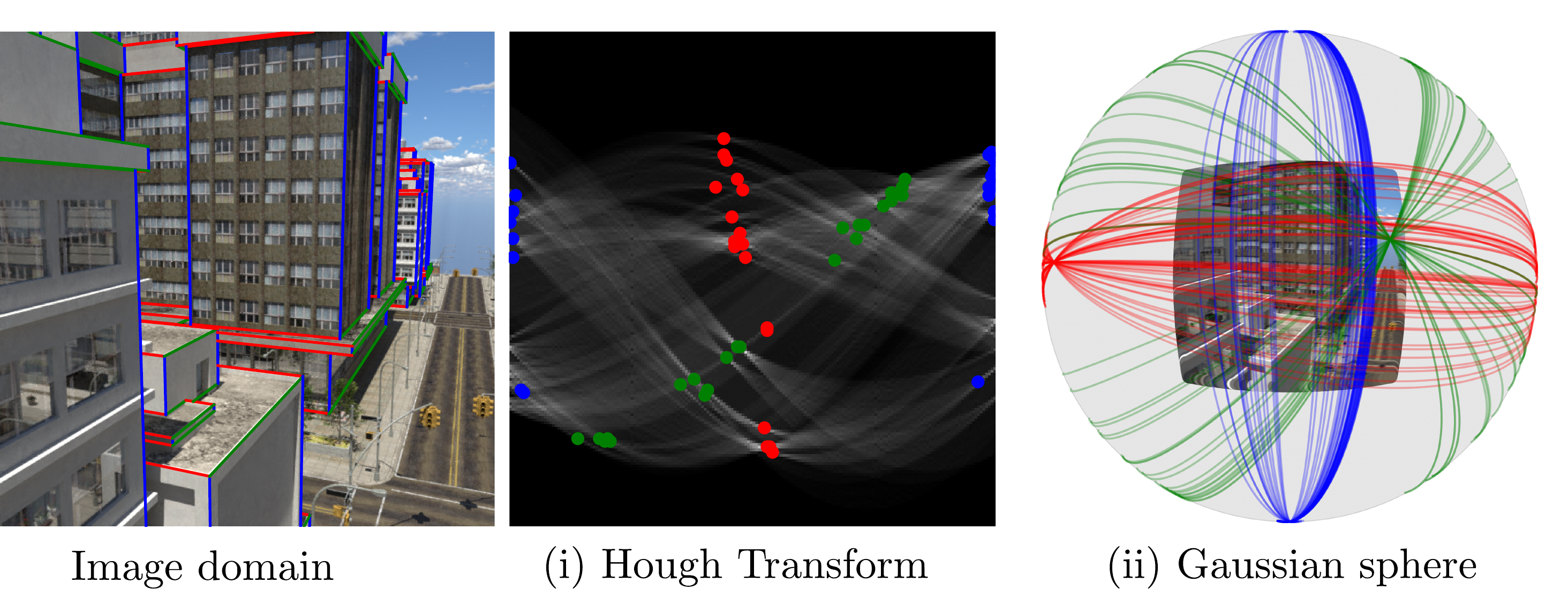}\\ 
    \end{tabular}
    \caption{\small
    We add two geometric priors: (i) Hough Transform and (ii) Gaussian sphere mapping, for vanishing points detection.
    We transform learned image features to the Hough domain, where lines are mapped to individual bins. 
    We further project the Hough bins to the Gaussian sphere, where lines become great circles and vanishing points are at the intersection of great circles. 
    Each color represents a set of image lines related to a vanishing point.
    Adding geometric prior knowledge makes our model data-efficient, less dependent on domain-specifics, and easily adaptable to problem variations such as detecting a variable number of vanishing points.
    }
    \label{fig:geometry}
\end{figure}

Vanishing point detection in images has non-vanishing real-world returns: camera calibration \cite{cipolla1999camera, antone2000automatic, grammatikopoulos2007automatic}, scene understanding \cite{flint2011manhattan}, visual SLAM \cite{davison2007monoslam, li2019leveraging}, or even autonomous driving \cite{lee2017vpgnet}. Deep learning is an excellent approach to vanishing point detection \cite{borji2016vanishing,chang2018deepvp,zhang2018dominant,zhou2019neurvps}, where all geometric knowledge is learned from large annotated data sets. Yet, in the real-world, there are several factors that complicate deep learning solutions: (1) Manually annotating large training sets is expensive and error prone; (2) Training models on large data sets require costly computational resources; (3) Practical changes to data collection cause domain shifts, hampering deep network generalization; (4) Slight changes in the problem setting require a complete change in deep network architectures. Thus, there is a need to make deep learning less reliant on data, and its architectures more robust to variants of the same problem.   

In this paper, we add geometric priors to deep vanishing point detection. Using geometric priors is data-efficient as this knowledge no longer needs to be learned from data. Thus, fewer annotations and compute resources are needed. Moreover, by relying on priors, the model is less sensitive to particular idiosyncrasies in the training data and generalizes better to domains with slightly different data distributions. Another advantage of a knowledge-based approach is that it is interpretable, and thus the architecture is easy to adapt to a slightly different problem formulation.

We add two geometric priors, see \fig{geometry}: (i) the Hough Transform and (ii) a Gaussian sphere mapping. 
Our trainable Hough Transform module represents each line as an (offset, angle) pair in line polar coordinates, allowing us to identify individual lines in Hough space \cite{duda1972use}. 
We subsequently map these lines from Hough space to the Gaussian sphere, where lines become great circles, and vanishing points are located at the intersection of great circles \cite{barnard1983interpreting}.
The benefit of using great circles is that  lines are mapped from the unbounded image plane to a bounded unit sphere, facilitating vanishing point detection outside the image view. 
Both the Hough Transform and the Gaussian sphere mapping are end-to-end trainable, taking advantage of learned representations, while adding knowledge priors.

This paper makes the following contributions: 
$(1)$ we add two geometric priors for vanishing point detection by mapping CNN features to the Hough Transform, and mapping Hough bins to the Gaussian sphere; 
$(2)$ we validate our choices and demonstrate similar accuracy as existing models on the large ScanNet \cite{dai2017scannet} and SceneCity Urban 3D \cite{zhou2019learning};
$(3)$ we show that adding prior knowledge increases data-efficiency, improving accuracy for smaller datasets;
$(4)$ we demonstrate our ability to tackle a different problem variant: detecting a varying numbers of vanishing points on the NYU Depth \cite{nathan2012indoor} dataset, where the number of vanishing points varies drastically from 1 to 8;
$(5)$ we show that adding prior knowledge reduces  domain shift sensitivity, which we validate by cross-dataset testing.

 \section{Related work}
 


\smallskip\noindent\textbf{Geometry-based vanishing point detection.}
Vanishing points occur at intersections of straight lines. Lines can be found by contour detection~\cite{zhou2017detecting} or a dual point-to-line mapping~\cite{lezama2014finding}. The common approach, however, is  using an explicit straight-line parameterization in the Hough Transform \cite{lutton1994contribution, quan1989determining, palmer1993optimised}. We exploit this straight-line parameterization as prior knowledge in a Hough Transform module.

Combining lines to vanishing points 
can be done by measuring the probability of a group of lines passing through the same point~\cite{tai1992vanishing}, voting schemes~\cite{gamba1996vanishing, wu2021real}, or hypothesis testing by counting the number of inlier lines such as J-Linkage \cite{toldo2008robust} used in \cite{feng2010semi, tardif2009non}. 
Other approaches see vanishing point detection as a grouping problem by applying line clustering 
\cite{barinova2010geometric, mclean1995vanishing,schaffalitzky2000planar}, expectation-maximization \cite{antone2000automatic,denis2008efficient,kovsecka2002video,schindler2004atlanta}, or branch-and-bound \cite{bazin2012globally, bazin2012globallyaccv, joo2018robust}. While these methods work well, they do not exploit prior knowledge of the 3D world.

A strong geometric prior for vanishing points is modeled by the Gaussian-sphere~\cite{barnard1983interpreting, collins1990vanishing}. A line in an image represents a great circle on the Gaussian-sphere and the intersections of great circles on the sphere denote vanishing points, detected as local maxima \cite{barnard1983interpreting, quan1989determining}. Mapping lines to the Gaussian sphere shifts the problem from the unbounded image plane to the constrained parameter space defined by the Gaussian sphere~\cite{collins1990vanishing,magee1984determining,straforini1993extraction}. Constraining the search space is a form of regularization, and is particularly beneficial for limited-data deep learning.  We exploit this prior knowledge by incorporating the Gaussian sphere mapping.


\smallskip\noindent\textbf{Learning-based vanishing point detection.} 
Vanishing point detection can be learned from large annotated datasets~\cite{borji2016vanishing,chang2018deepvp,zhai2016detecting,zhang2018dominant}. 
It is effective to split the problem in separate stages: line detection, inverse gnomonic projection, network training and post-processing, as in Kluger~\etal~\cite{kluger2017deep}. 
Conic convolutions on hemisphere points, provided further improvements, in Zhou~\etal~\cite{zhou2019neurvps}. 
In contrast, rather than focusing on accurate large-scale deep models, we consider challenging real-world scenarios, such as: limited training samples, cross-dataset domain switch, and non-Manhattan world.

\smallskip\noindent\textbf{Robustness to domain shifts.} 
Classical solutions to vanishing point detection~\cite{li2020quasi, feng2010semi, simon2018contrario} are built exclusively on prior-knowledge. Such methods are data-free, and thus designed to work on any domain. Yet, they cannot take advantage of expressive deep-feature learning for vanishing point detection~\cite{ zhou2019neurvps, liu2021vapid}. On the other hand, deep models are notoriously sensitive to distribution shifts between training and test~\cite{luo2019taking, lengyel2021zero}. Active research on this includes: domain adaptation~\cite{wang2018deep, wulfmeier2017addressing, peng2018zero}, domain generalization~\cite{zhou2021domain}, multi-domain learning~\cite{rebuffi2018efficient, li2019efficient}, etc. Such solutions entail significant changes to the deep network model, adding complexity for practical real-world applications. 
Hence, we focus on a single method which does not requiring large model changes for robustness to minor domain shifts. Our goal is to combine the robustness of knowledge-based methods, with the power of deep representation learning.

\smallskip\noindent\textbf{Manhattan versus non-Manhattan world.}
The Manhattan world assumes exactly 3 vanishing points. This assumption has been proven useful for orthogonal vanishing point detection \cite{antunes2013global,bazin2012globally,mirzaei2011optimal,wildenauer2012robust}. However, the Manhattan assumption does not hold in several real-world scenarios such as non-orthogonal walls and wireframes in man-made structures. Vanishing point detection in non-Manhattan world is done by robust multi-model fitting \cite{kluger2020consac}, horizon line detection~\cite{zhai2016detecting}, branch-and-bound with a novel mine-and-stab strategy \cite{li2020globally}, Bingham mixture model fitting \cite{li2021learning} or non-maximum suppression on the Gaussian sphere \cite{liu2021vapid}. In our work, we refrain from adding explicit orthogonality constraints, which makes our method applicable to non-Manhattan scenarios as well. We rely on the Hough Transform and the Gaussian sphere to map pixel-wise representations to the entire hemisphere. And using a clustering algorithm we detect multiple vanishing points simultaneously.

\section{Geometric priors for VP detection}
 \begin{figure*}[t!]
    \centering
    \includegraphics[width=0.9\textwidth]{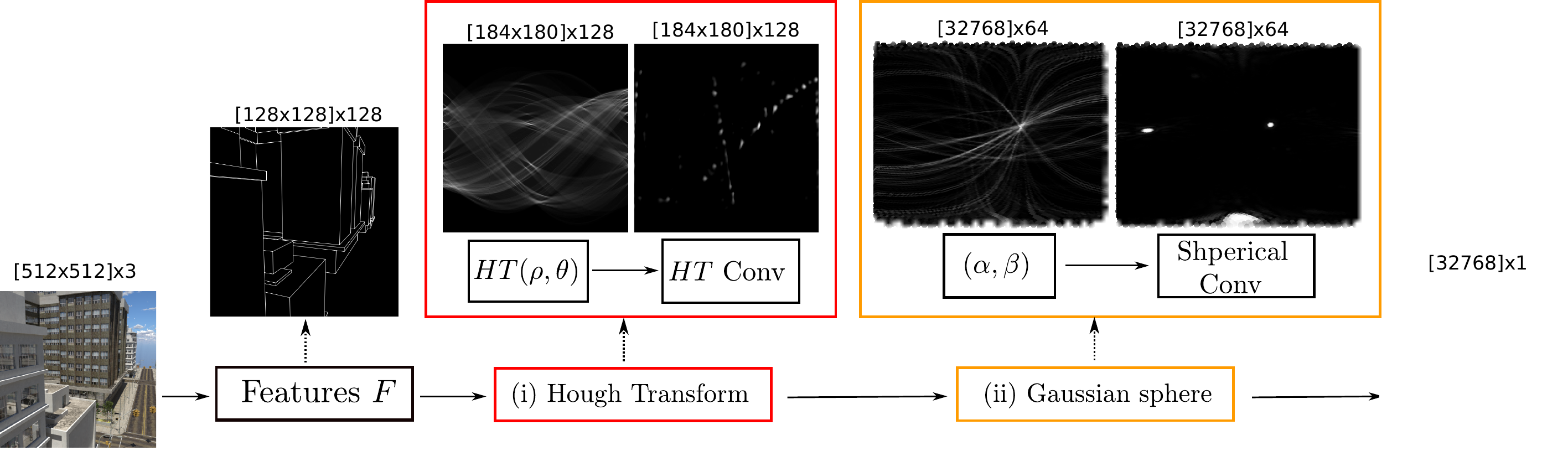}
    \caption{\small
    \textbf{Overview:}
    The model starts from in input image, and predicts vanishing points on the Gaussian hemisphere by relying on two geometric priors: (i) Hough Transform, and (ii) Gaussian sphere mapping.
    We use a convolutional network to learn features which are then mapped to Hough space, where each bin is a line. 
    We filter the Hough space and project Hough bins to the Gaussian hemisphere and apply spherical convolutions to find vanishing points.
    We indicate the size of the learned features above, where the last dimension is the number of channels. 
    We sample 32,768 points on the hemisphere using the Fibonacci lattice \cite{gonzalez2010measurement}, resulting in features maps of size 32,768. Our model learns to classify spherical points as vanishing points or not using a binary cross-entropy loss. There is \emph{no} intermediate supervision. }
    \label{fig:overview}
\end{figure*}

\noindent\textbf{General outline of our approach.}  
\fig{overview} depicts the overall structure of our model.  
We build on two geometric priors: (i) Hough Transform, and (ii) Gaussian sphere mapping. 
A CNN learns image features, which are then mapped to a line parameterization via Hough Transform. 
We project the features of parameterized lines to the Gaussian sphere where spherical convolutions precisely localize vanishing points. 

\medskip\noindent\textbf{ (i) Hough Transform. }
Similar to \cite{zhou2019neurvps}, we use a single-stack hourglass network \cite{newell2016stacked} to extract image features, $F$ to be mapped into Hough space~\cite{lin2020deep}, $HT$.  
The $HT$ space parameterizes image lines in polar coordinates using a set of discrete offsets $\rho$ and discrete angles $\theta$, defining a 2D discrete histogram.
In practice, a set of pixels $(x(i), y(i))$ along a line indexed by $i$, vote for a line parameterization to which they all belong:  
\begin{align}
    HT(\rho, \theta)  =  \sum_i F(\rho \cos \theta - i \sin \theta, \rho \sin \theta + i \cos \theta)    
\end{align}
The Hough Transform module starts from an $[H {\times} W]$ feature map $F$ and outputs an $[N_{\rho} {\times } N_{\theta}]$ Hough histogram $HT$, where $N_{\rho}$ and $N_{\theta}$ are the number of sampled offsets and angles in Hough Transform. 
We set $H{=}128$, $W{=}128$, $N_{\rho}{=}184$, and $N_{\theta}{=}180$.  
This results in $[N_{\rho} {\times} N_{\theta}]$ possible line parameterizations. 
We find the local maxima in the Hough domain by performing a 1D convolutions over the offsets. This removes the noisy responses in the Hough space, as in ~\fig{overview}. We refer the readers to \cite{lin2020deep} for details.

\begin{figure*}[t!]
    \centering
    \vspace{-10px}
    \begin{tabular}{c@{\hskip 1.0in}c}
        \includegraphics[width=0.4\textwidth]{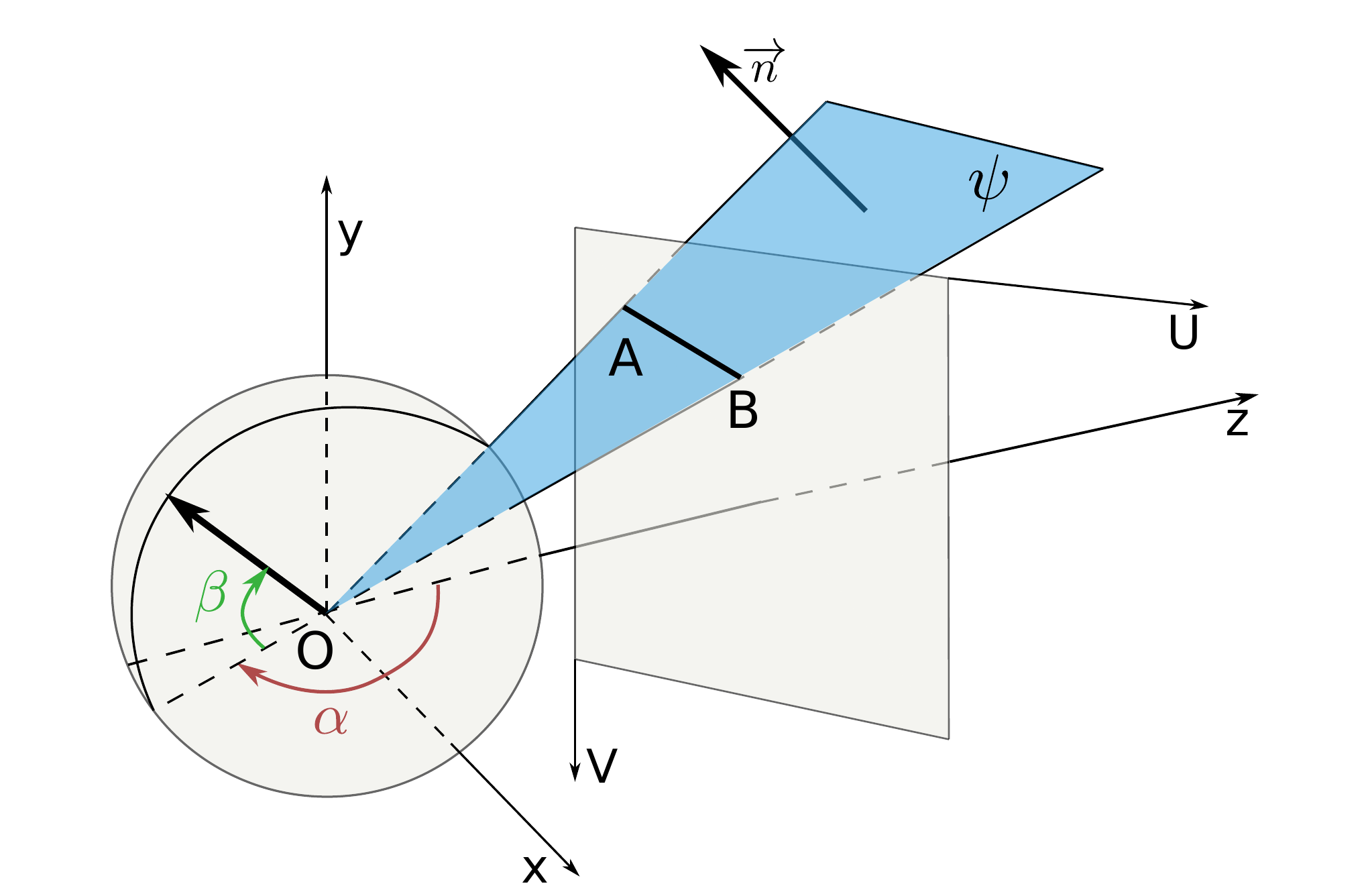}& 
        \includegraphics[width=0.4\textwidth]{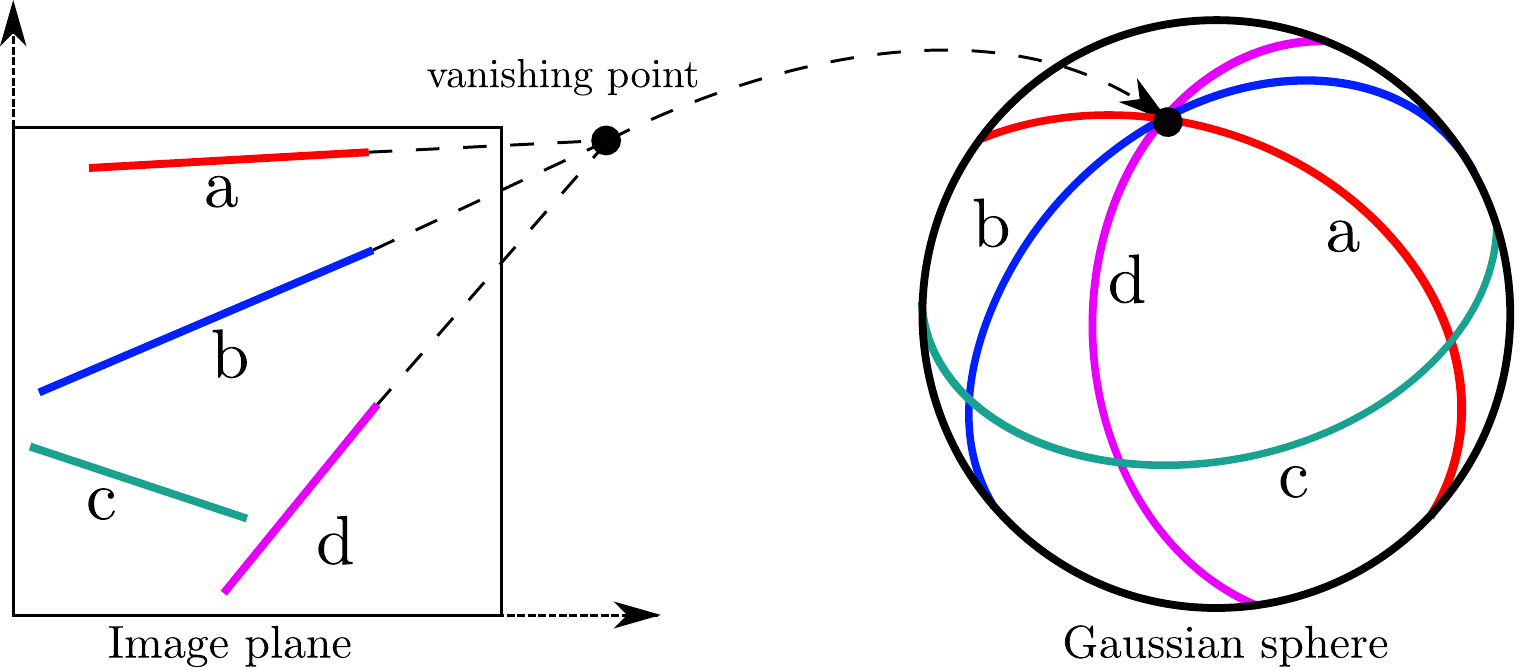} \\
        (a) The Gaussian sphere & (b) Vanishing points on the Gaussian sphere\\
    \end{tabular}
    \caption{\small
    \textbf{Gaussian sphere representations for vanishing points \cite{barnard1983interpreting, quan1989determining}:} 
    (a) The Gaussian sphere is a unit sphere located at the camera center, $\mathbf{O}$. 
    Points on the sphere are encoded by two angles: $(\alpha, \beta)$ the azimuth and the elevation, respectively.
    A line segment $\mathbf{AB}$ in the image plane together with the camera center $\mathbf{O}$ forms a plane $\psi$, highlighted in blue.
    To define the mapping from the image to the sphere, we only need to know the normal $\protect \overrightarrow{n}$ to the plane $\psi$.
    (b) Image lines are projected as great circles on the sphere. 
    The intersection of multiple great circles on the sphere represents a vanishing point.
    }
    \label{fig:gaussian_sphere}
\end{figure*}

\medskip\noindent\textbf{ (ii.1) Gaussian sphere mapping.}  The Gaussian sphere is a unit sphere centered at the camera origin, $\mathbf{O}$. Vanishing points on the sphere are represented as normalized 3D line directions $\delta$.

Starting from a bin in the Hough domain $(\rho_{AB}, \theta_{AB})$, corresponding to a line direction in the image plane $\overrightarrow{AB}$, we want to map this to the Gaussian sphere.
Two image points $\mathbf{A}$ and $\mathbf{B}$ sampled from a line represented by its $HT$ bin $(\rho_{AB}, \theta_{AB})$, together with the camera center $\mathbf{O}$, form a plane $\psi$ as depicted in~\fig{gaussian_sphere}(a).
The plane $\psi$ is described by its normal vector: 
\begin{equation}
\overrightarrow{n} = (n_{x}, n_{y}, n_{z}) = \frac{\overrightarrow{OA} \times \overrightarrow{OB}}{\norm{\overrightarrow{OA} \times \overrightarrow{OB}}}.
\end{equation}
This normal vector $\overrightarrow{n}$ is the only information we need to map the image line direction $\overrightarrow{AB}$ to the Gaussian sphere.

The spherical coordinates $(\alpha, \beta)$ describe a point on the Gaussian sphere, where 
$\alpha$ is the azimuth defined as the angle from the $z$-axis in the $xz$ plane, and
$\beta$ is the elevation representing the angle measured from the $xz$ plane towards the $y$-axis, as shown in~\fig{gaussian_sphere}(a). 
The intersection between the plane $\psi$ and the Gaussian sphere is a great circle. 
This great circle represents the projection of the image line direction $\overrightarrow{AB}$ on the Gaussian sphere. Intersections of multiple great circles are potential vanishing points, see ~\fig{gaussian_sphere}(b). 

We compute the projection of the image line direction $\overrightarrow{AB}$, by estimating the elevation $\beta$ as a function of the azimuth $\alpha$ and the normal vector $\overrightarrow{n}$ \cite{quan1989determining}:
\begin{alignat}{1}
    \beta(\alpha, \overrightarrow{n}) &= \tan^{-1}  \frac{-n_{x} \sin{\alpha} - n_{z} \cos{\alpha}}{n_{y}},
    \label{eq:gaussian_mapping}
\end{alignat} 
where we uniformly sample $\alpha$ in the range $[-\pi/2, \pi)$.

Because the Gaussian sphere is symmetric we only need a hemisphere. We sample $N$ points on the Gaussian hemisphere using a Fibonacci lattice \cite{gonzalez2010measurement}  and then project lines, corresponding to bins in the Hough space, to these $N$ sampled sphere points. 
For each line parameterization in Hough space $(\rho,\theta)$, we first compute its normal vector $\overrightarrow{n}$.
We, then, estimate its corresponding $(\alpha, \beta)$ spherical coordinates using~\eq{gaussian_mapping}. 
We subsequently assign each $(\alpha, \beta)$ pair to its nearest neighbor in the sampled points from the Fibonacci lattice, by computing their cosine distance. 
To parallelize this process, we precompute the projection of all Hough line parameterizations onto the sampled sphere locations. 
This mapping is stored in an $[N_{\rho} {\times} N_{\theta} {\times} M]$ tensor, where $[N_{\rho} {\times} N_{\theta}]$ is the number of line parameterizations in Hough space and $M$ is the number of sampled azimuth angles $\alpha$. 
We set $N$=32,768 and $M$=1,024.

\medskip\noindent\textbf{ (ii.2) Spherical convolutions on the hemisphere.} 
We employ spherical convolutions to predict vanishing points. 
We treat the points sampled on the hemisphere as a point cloud and use EdgeConv~\cite{wang2019dynamic} to convolve over the hemisphere. EdgeConv operates on a k-nearest neighbor graph on the points. It learns to represent local neighborhoods by applying a non-linear function to the neighbors' features, and then aggregates those features with a symmetric operator.
The neighbors' features are localized by subtracting the features of the centroid.
Like \cite{wang2019dynamic}, we take the per-feature maximum over the neighbors to aggregate edge features.

\begin{figure}[t!]
    \centering
    \includegraphics[width=0.5\textwidth]{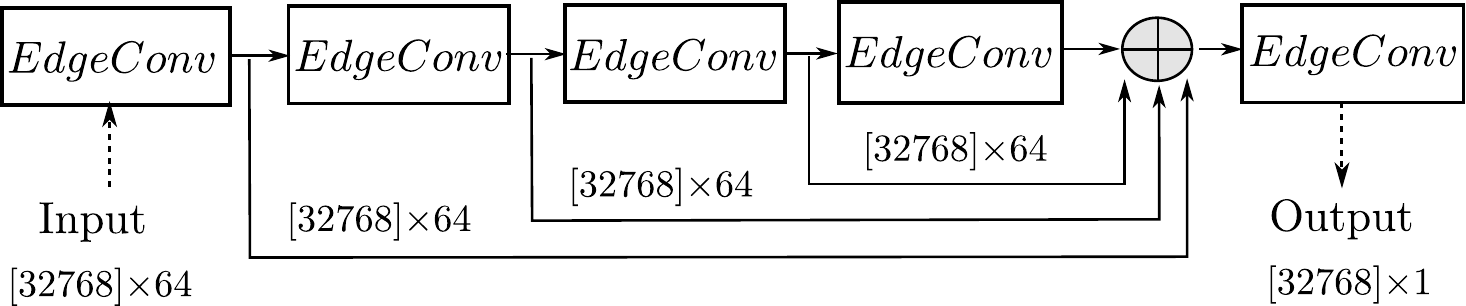}
    \caption{\small
    \textbf{Spherical convolutions on the hemisphere.} 
    We use EdgeConv \cite{wang2019dynamic} for precise vanishing point localization on the Gaussian sphere. The concatenation of the previous feature maps is fed into the final layer to produce a prediction.}
    \label{fig:sphere_conv}
\end{figure}

As shown in \fig{sphere_conv}, the spherical part of our model contains 5 EdgeConv modules \cite{wang2019dynamic}. Each EdgeConv module transforms neighboring features with a fully connected layer, a BatchNorm layer \cite{ioffe2015batch} and a LeakyReLU activation.
We use $N$=32,768 nodes on the hemisphere and compute the 16 nearest neighbors for each node. We concatenate the features maps from previous layers and feed them into the last EdgeConv layer to produce the final prediction. 

\medskip\noindent\textbf{Model training and inference.} \label{training} 
We train the model using the binary cross-entropy loss. For each annotated vanishing point, we label its nearest neighbor in the sampled points as $+1$ and the others as $0$.
Because the number of positive samples is considerably lower than the negative samples, we compute two separate average losses over the positive and the negative samples, and then sum these. 
There is \emph{no} intermediate supervision or guidance.

During inference, we use DBSCAN \cite{ester1996density, pedregosa2011scikit} to cluster all points on the Gaussian sphere based on the cosine distance. 
The \emph{eps} parameter of DBSCAN  \cite{pedregosa2011scikit} is set to be 0.005. 
The point with the highest confidence in each cluster is the prediction. 
We rank all predictions by confidence.

\begin{figure}[t!]
    \centering
    \includegraphics[width=0.45\textwidth]{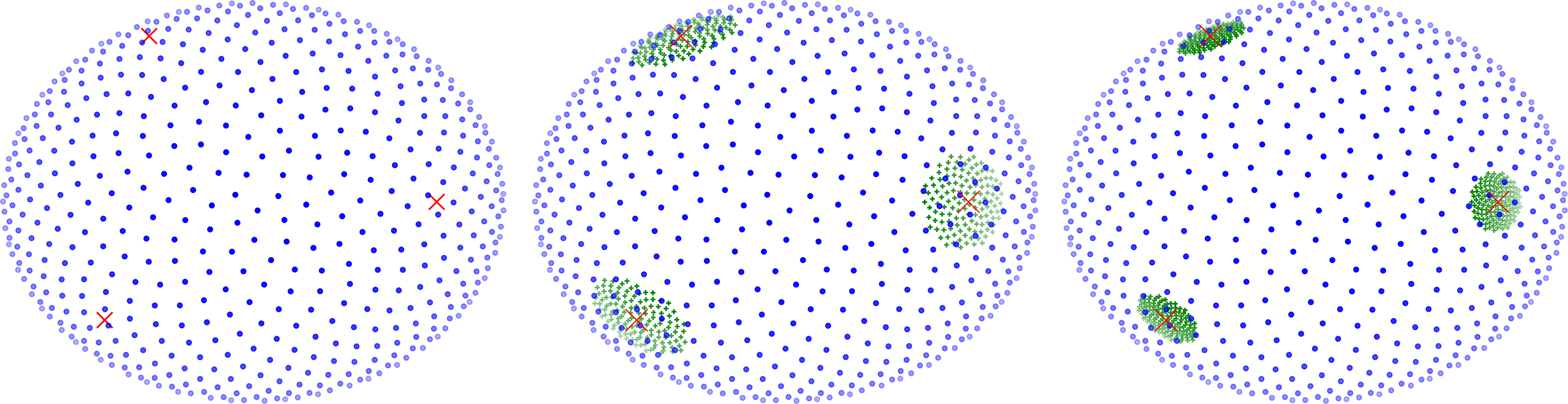}
    \caption{\small
    \textbf{Multi-scale sampling on the hemisphere.} 
    We sample points at three scales for detecting vanishing points in the Manhattan world, as in \cite{zhou2019neurvps}. 
    Blue indicates sampling at the first scale, while green indicates fine-grained sampling at the following scales. The red crosses are the predictions at each scale.
    }
    \label{fig:sampling_multi}
\end{figure}

\medskip\noindent\textbf{Multi-scale sampling with the Manhattan assumption.} \label{sampling_multi}
For the Manhattan world, we know beforehand that there are only 3 orthogonal vanishing points, therefore in this scenario, we can use a multi-scale sampling strategy to reduce computation, as in \cite{zhou2019neurvps}. Here, we sample points and apply spherical convolutions at 3 scales: $\delta\approx\{90^{\circ}, 13^{\circ}, 4^{\circ}\}$ and $N=\{512, 128, 128\}$, where $\delta$ controls the sampling radius and $N$ indicates the number of sampled points respectively. \fig{sampling_multi} displays the multi-scale sampling.  
The spherical convolution networks share the same architecture while processing different number of samples. We provide details in the supplementary material.

\section{Experiments}

\noindent\textbf{Datasets.} 
We evaluate on three datasets following the Manhattan world assumption: SU3 (SceneCity Urban 3D) \cite{zhou2019learning}, ScanNet \cite{dai2017scannet}, YUD \cite{denis2008efficient}, as well as the NYU Depth \cite{nathan2012indoor} dataset which does not follow the Manhattan world assumption. The SU3 dataset contains $23$K synthetic images, which are split into $80\%$, $10\%$ and $10\%$ for training, validation and testing respectively. The ScanNet has more than 200K real-world images, among which 189,916 examples are used for training. The ``ground truth" VPs are estimated from surface normals as in \cite{zhou2019neurvps}, thus being less precise than other datasets.
In the NYU Depth dataset, the number of vanishing points varies from 1 to 8 across images, making it more challenging. 
The NYU Depth dataset has 1,449 images, approximately $\times 200$ and $ \times 20$ smaller than the ScanNet dataset and the SU3 dataset, respectively, further increasing the difficulty of training CNN models.
We additionally demonstrate the effect of geometric priors on the small-scale YUD dataset with only 102 images. Detailed comparisons are in the supplementary material. 
Unless specified otherwise, we use the ground-truth focal length on SU3, ScanNet and YUD for the Manhattan assumption.

\smallskip\noindent\textbf{Evaluation.} 
On the SU3, ScanNet and YUD datasets (Manhattan assumption), we evaluate the angle difference between the predicted and the ground-truth vanishing points in the camera space, as in \cite{zhou2019neurvps, kluger2020consac, liu2021vapid}. 
We then estimate the percentage of the predictions that have a smaller angle difference than a given threshold and compare the angle accuracy (AA) under different thresholds, as in \cite{zhou2019neurvps, liu2021vapid}. We use the ground-truth focal length to exploit the orthogonal constraint.
On the NYU Depth dataset we follow \cite{kluger2020consac} and first rank detected vanishing points by confidence, and then use the bipartite matching \cite{crouse2016assignment} to calculate the angular errors for the top $k$ predictions. After matching, we generate the recall curve and measure the area under the curve (AUC) up to a threshold, e.g. $10^{\circ}$. 

\begin{figure}[t!]
    \centering
    \includegraphics[width=0.4\textwidth]{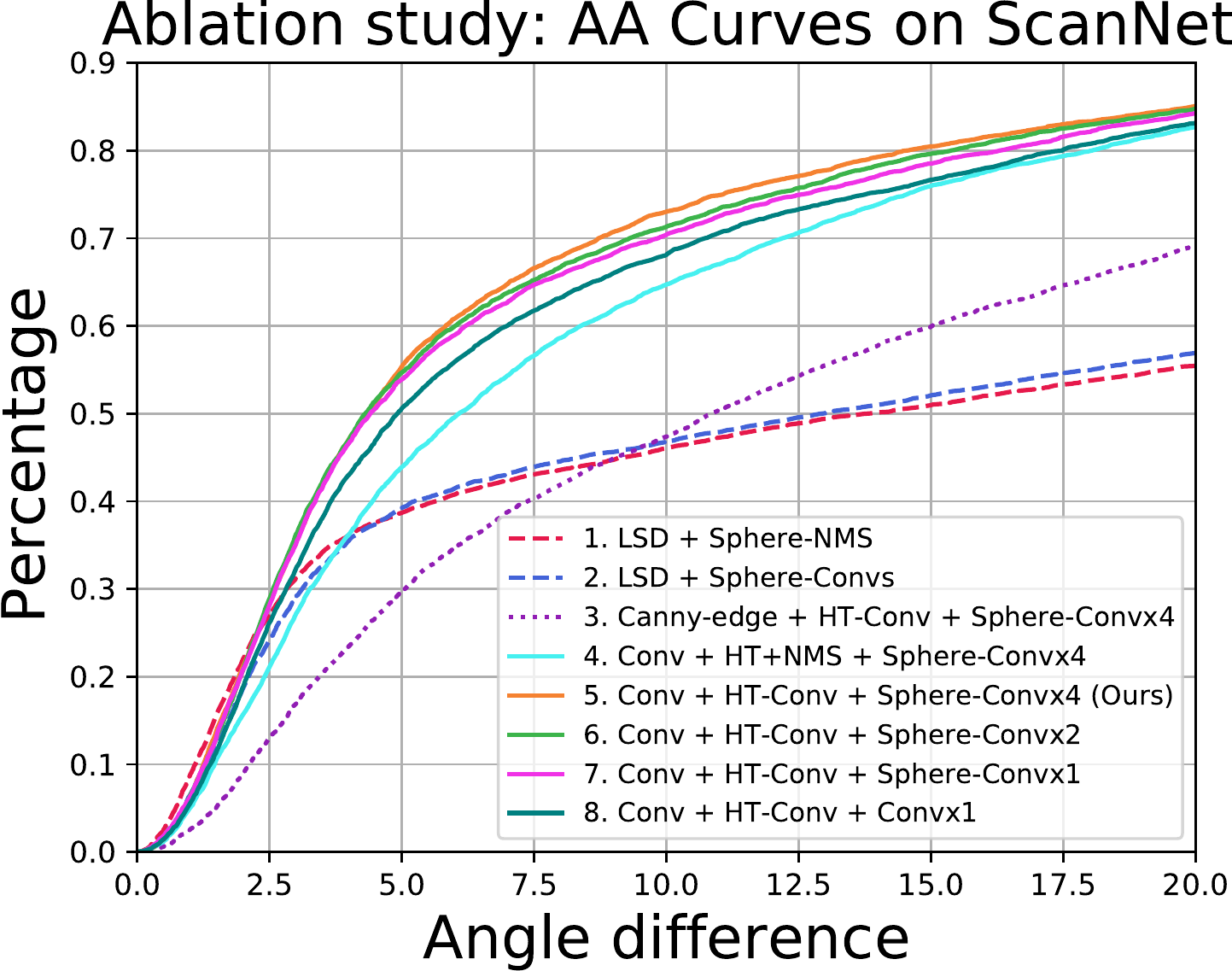}\\
    \caption{\small
    \textbf{Exp 1: Model choices.}  We show the effects of the two geometric priors quantitatively on the ScanNet-$1\%$ subset. Adding HT layers and spherical convolutions outperforms the baselines, thus demonstrating the effectiveness of geometric priors. 
    }
\label{fig:ablation_scannet}
\vspace*{-10px}
\end{figure}

\begin{table}[tbp!]
    \centering
        \resizebox{0.8\linewidth}{!}{
        \begin{tabular}{cc@{\hskip 0.2in}c@{\hskip 0.2in}c@{\hskip 0.2in}c@{\hskip 0.2in}c@{\hskip 0.2in}}
        \toprule
        & \multicolumn{2}{c}{HT ($\#$ of angles)} & \multicolumn{3}{c}{Sphere ($\#$ of points)} \\
       \cmidrule(l){2-3}  \cmidrule(l){4-6}
        & $90$ & $180$ & $8K$ & $16K$ & $32K$ \\  \midrule
        AA$@3^{\circ}$   & 77.1 &  79.3 & 73.3 & 77.4 & 79.3\\
        \bottomrule
        \end{tabular} 
        }
    \caption{\textbf{Quantization analysis on SU3-$10\%$ subset.} Denser samplings improve performance. In practice, we uniformly sample 180 angles from $[0, \pi)$ for HT, and $32K$ points on the sphere.}
    \label{tab:numerical_sampling}
    \vspace{-10px}
\end{table}

\smallskip\noindent\textbf{Baselines.} 
We compare our model with J-Linkage \cite{feng2010semi}, Contrario-VP \cite{simon2018contrario}, Quasi-VP \cite{li2019quasi, li2020quasi},  NeurVPS \cite{zhou2019neurvps}  and CONSAC \cite{kluger2020consac} on SU3, ScanNet and YUD. 
On the non-Manhattan NYU Depth dataset we only compare with J-Linkage, T-Linkage \cite{magri2014t}, CONSAC and VaPid \cite{liu2021vapid}, as the other models rely on the Manhattan assumption. 
J/T-Linkage, Contrario-VP and Quasi-VP are non-learning methods, employing line segment detection \cite{von2008lsd}. 
NeurVPS and our model are end-to-end trainable, while CONSAC needs line segments as inputs. We follow the official implementations and use the default hyperparameters to reproduce all results. 
We do not consider the baselines \cite{wu2021real, li2021learning, liu2021vapid} due the lack of code\slash results on certain datasets.

\smallskip\noindent\textbf{Implementation details.} 
We implement our model in Pytorch \cite{paszke2017automatic}, and provide the code online \footnote{\url{https://github.com/yanconglin/VanishingPoint_HoughTransform_GaussianSphere}}.
Our models are trained from scratch on Nvidia RTX2080Ti GPUs with the Adam optimizer \cite{kingma2014adam}. 
The learning rate and weight decay are set to be $4 \times 10^{-4}$ and $1 \times 10^{-5}$, respectively. To maximize GPU usage, we set the batch size to 4 and 16 when using multi-scale sampling. 
On the SU3 and NYU Depth datasets, we train the model for a maximum of 36 epochs, with the learning rate decreases by 10 after 24 training epochs. 
On the ScanNet dataset, we train for 10 epochs and decay the learning rate by 10 after 4 epochs. On the YUD datset we use pre-trained models on SU3.

\begin{table*}[t!]
    \centering
    \resizebox{1\linewidth}{!}{
        \begin{tabular}{p{7em} r@{\hskip 0.2in}rc c cc ccc ccc}
        \toprule
        Datasets & & & \multicolumn{2}{c}{SU3 \cite{zhou2019learning}} & \multicolumn{3}{c}{ScanNet \cite{dai2017scannet}}& \multicolumn{3}{c}{YUD \cite{denis2008efficient}}\\ 
        \cmidrule(l){4-5} \cmidrule(l){6-8}\cmidrule(l){9-11}
        Metrics & Params & FPS & AA$@3^{\circ}$ & AA$@5^{\circ}$ &   AA$@3^{\circ}$  & AA$@5^{\circ}$ & AA$@10^{\circ}$ & AA$@3^{\circ}$  & AA$@5^{\circ}$ & AA$@10^{\circ}$\\  \midrule
        J-Linkage$^{\dagger}$ \cite{feng2010semi}  & --- & 1.0 & 82.0  & 87.2  & 15.7  & 27.3  & 43.0  & 60.8 & 71.8 & 81.5 \\
        \small{Contrario-VP$^{\dagger}$} \cite{simon2018contrario} & ---& 0.6 & 64.8  & 72.2 &12.0 & 21.4  &35.3   & 58.6 &70.7 & 81.8\\
        Quasi-VP \cite{li2019quasi} & --- & \textbf{29.0} & 75.9 & 80.7  &14.7 & 25.3 & 39.4  & 58.6 & 61.0 & 74.0\\
        CONSAC$^{\dagger}$ \cite{kluger2020consac} & 0.2 M &3.0  & 86.3 & 90.3  &15.8 & 24.6 & 36.0  & \textbf{61.7} & 73.6 & 84.4 \\
        NeurVPS \cite{zhou2019neurvps} & 22 M & 0.5 & \textbf{93.9} & \textbf{96.3} & 24.0 & 41.8 & \textbf{64.4}  & 52.4 & 64.0 & 77.8  \\
        \emph{Ours} & 7 M & 5.5 &84.0 & 90.2  & \textbf{24.8} & \textbf{42.1} & 63.7  & 60.7 &\textbf{74.3}  & \textbf{86.3}   \\
        \emph{Ours$^\ast$} & 5 M & 23.0 &84.8 & 90.7 & 22.9  & 39.8  & 62.4 & 59.5  &72.6  & 85.4   \\
        \emph{Ours$^{\dagger}$} & 7 M & 5.5 &81.7 & 88.7  & 22.2 & 38.8 & 59.9  & 59.1 & 72.6  & 84.6   \\
        \bottomrule
        \end{tabular} 
    } 
    \caption{\small
    \textbf{Exp 2: Manhattan world.} 
    Angular accuracy on SU3, ScanNet and YUD datasets. \emph{Ours} achieves the best results on the the YUD dataset, and is competitive on the larger ScanNet and SU3 datasets. \emph{Ours$^\ast$} adopts the multi-scale sampling strategy, thus being significantly faster. $^{\dagger}$ assumes unknown focal length, thus making the Manhattan assumption no longer applicable. \emph{Ours$^{\dagger}$} shows a constant decrease over \emph{Ours} across datasets, indicating the usefulness of the orthogonal constraint.  
    Supplementary material provides qualitative visualizations. 
    We conclude that adding priors does not reduce accuracy in the large scale setting. 
    }
    \label{tab:numerical_eval}
\end{table*}
\begin{figure*}[t!]
    \centering
    \begin{tabular}{ccc}
    \multicolumn{3}{c}{\includegraphics[width=0.8\textwidth]{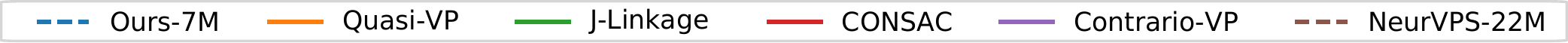}}\\
    \includegraphics[width=0.3\textwidth]{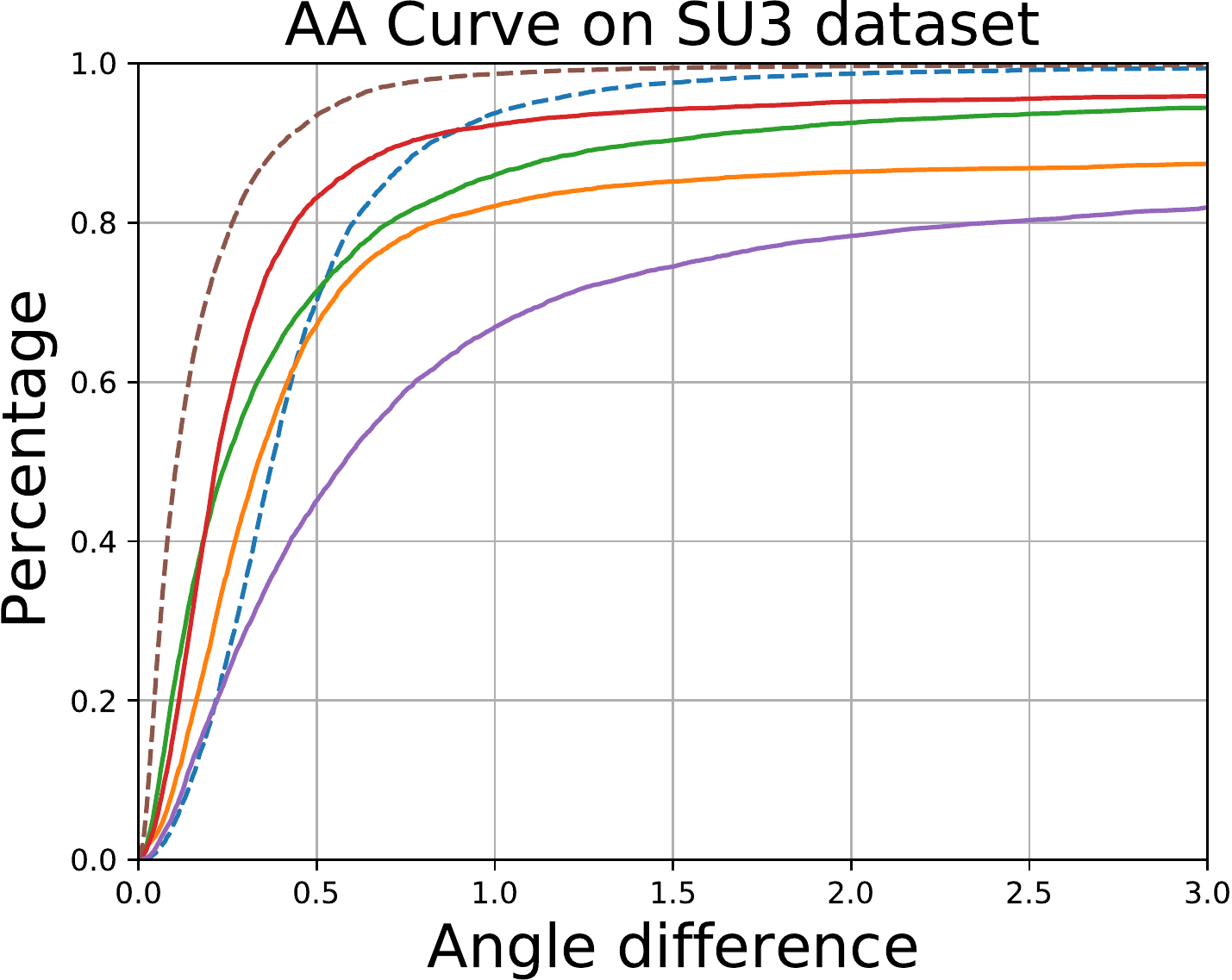}& 
     \includegraphics[width=0.3\textwidth]{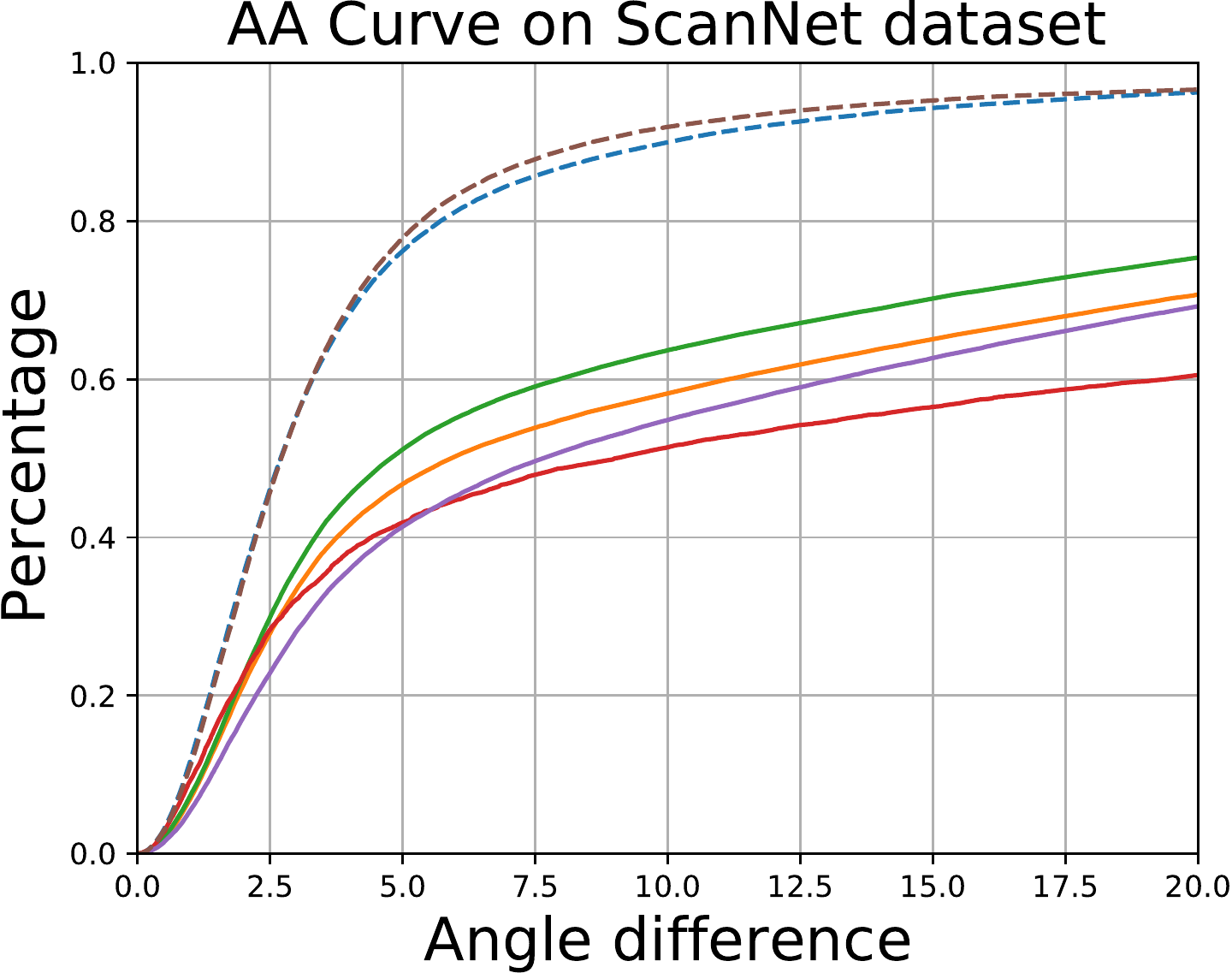}&
     \includegraphics[width=0.3\textwidth]{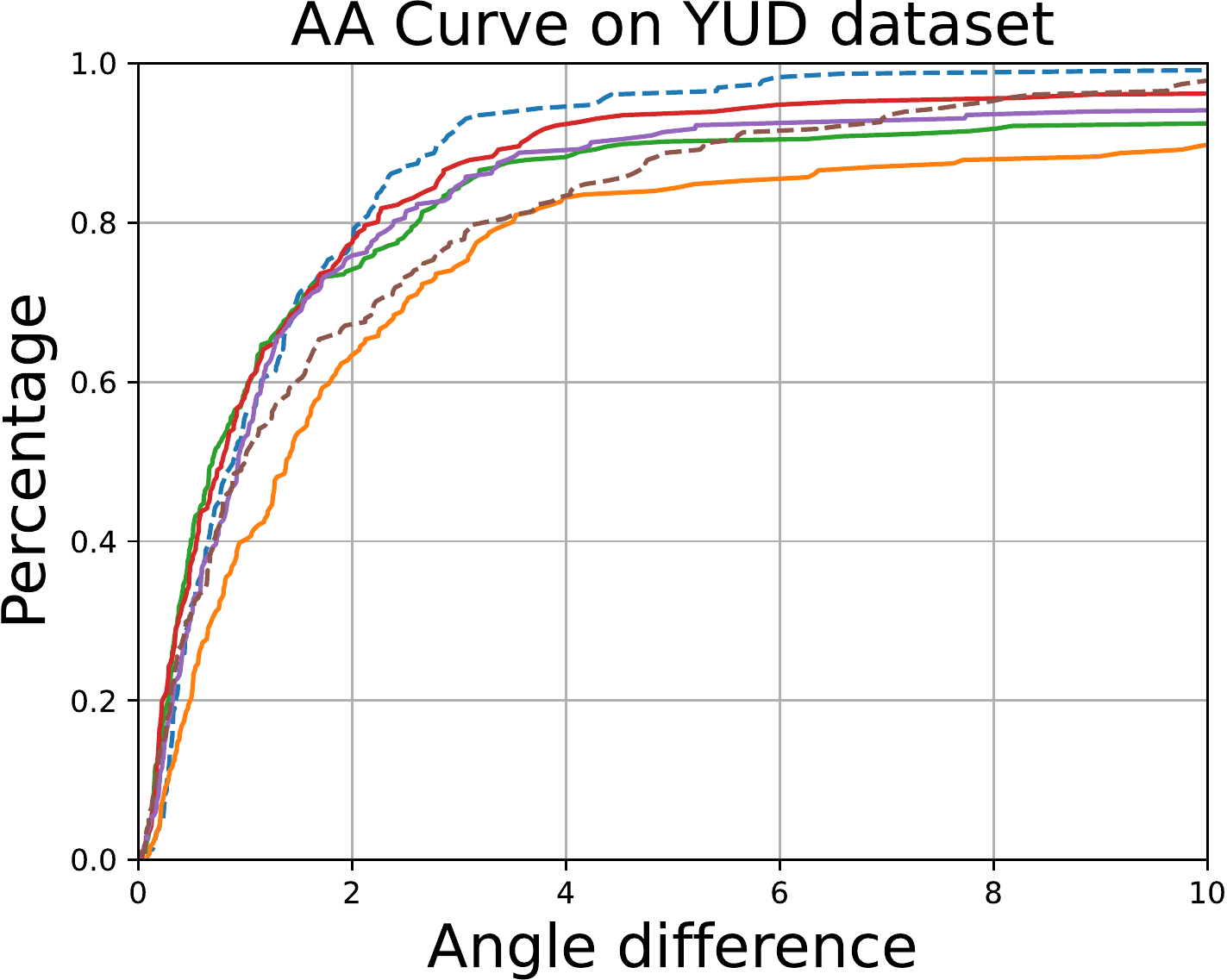} \\
    \end{tabular}
    \caption{\small
    \textbf{Exp 2: Manhattan world.}
     AA curves on on the ScanNet, SU3 and YUD datasets containing 3 orthogonal vanishing points.  Learning-based  approaches outperform methods relying purely on line segments and grouping, validating the power of representation learning.
     Our model shows comparable results to the best performing NeurVPS on ScanNet, while using 3$\times$ less parameters.  
     On the smaller YUD dataset, our model slightly exceeds state-of-the-art. Generally, with ample data, our approach is comparable to others. 
     }
    \label{fig:comparison_all}
\end{figure*}
\subsection{\textbf{Exp 1:} Evaluating model choices}
We evaluate on a subset of ScanNet containing $1\%$ of the data, and provide the results in \fig{ablation_scannet}. 
Model \emph{(1)} is a non-learning baseline using a classic line segment detector (LSD) \cite{von2008lsd} and non-maximum suppression (NMS) on the sphere. 
Model \emph{(2)} replaces the NMS with spherical convolutions, but still shows inferior result as LSD fails to detect reliable line segments. 
Model \emph{(3)} combines a Canny-edge detector, Hough Transform and spherical convolutions. 
Comparing \emph{(3-5)} indicates the added value of learning semantics from images, rather than using classic edge detectors. 
Comparing \emph{(4-5)} shows the effectiveness of backpropagating through Hough Transform. 
Comparing \emph{(5-8)} exemplifies the added value of spherical convolutions.
Our method combines both classical and deep learning approaches into an end-to-end trainable model.

We also evaluate the impact of quantizations numerically on the synthetic SU3-$10\%$ subset, which contains precise VP annotations, thus making quantization a crucial factor. As shown in \tab{numerical_sampling}, fine-grained sampling is essential for a better result. 

\begin{figure*}[t]
    \centering
    \begin{tabular}{c@{\hskip 1.0in}c}
        \includegraphics[width=0.40\textwidth]{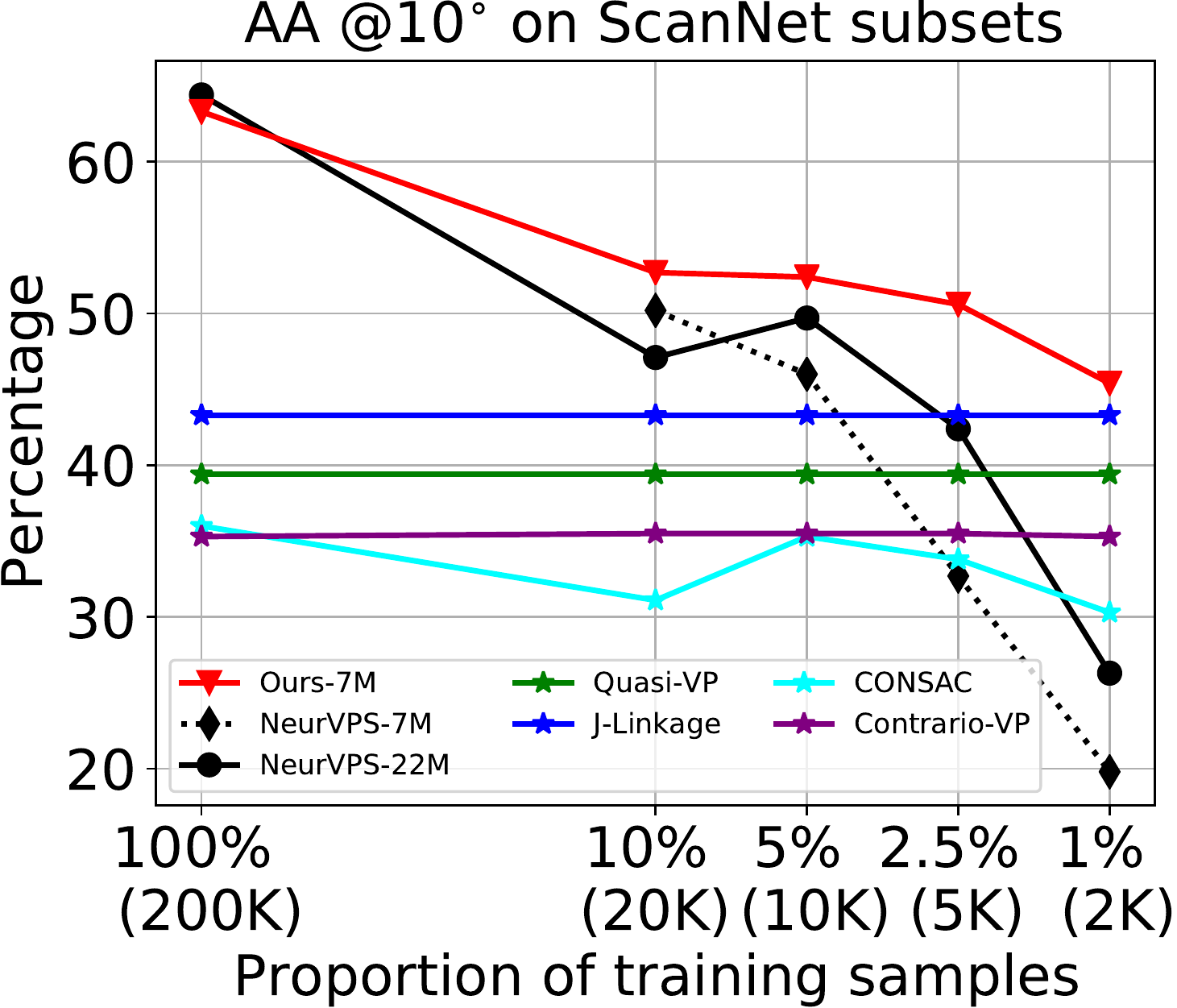}& 
        \includegraphics[width=0.42\textwidth]{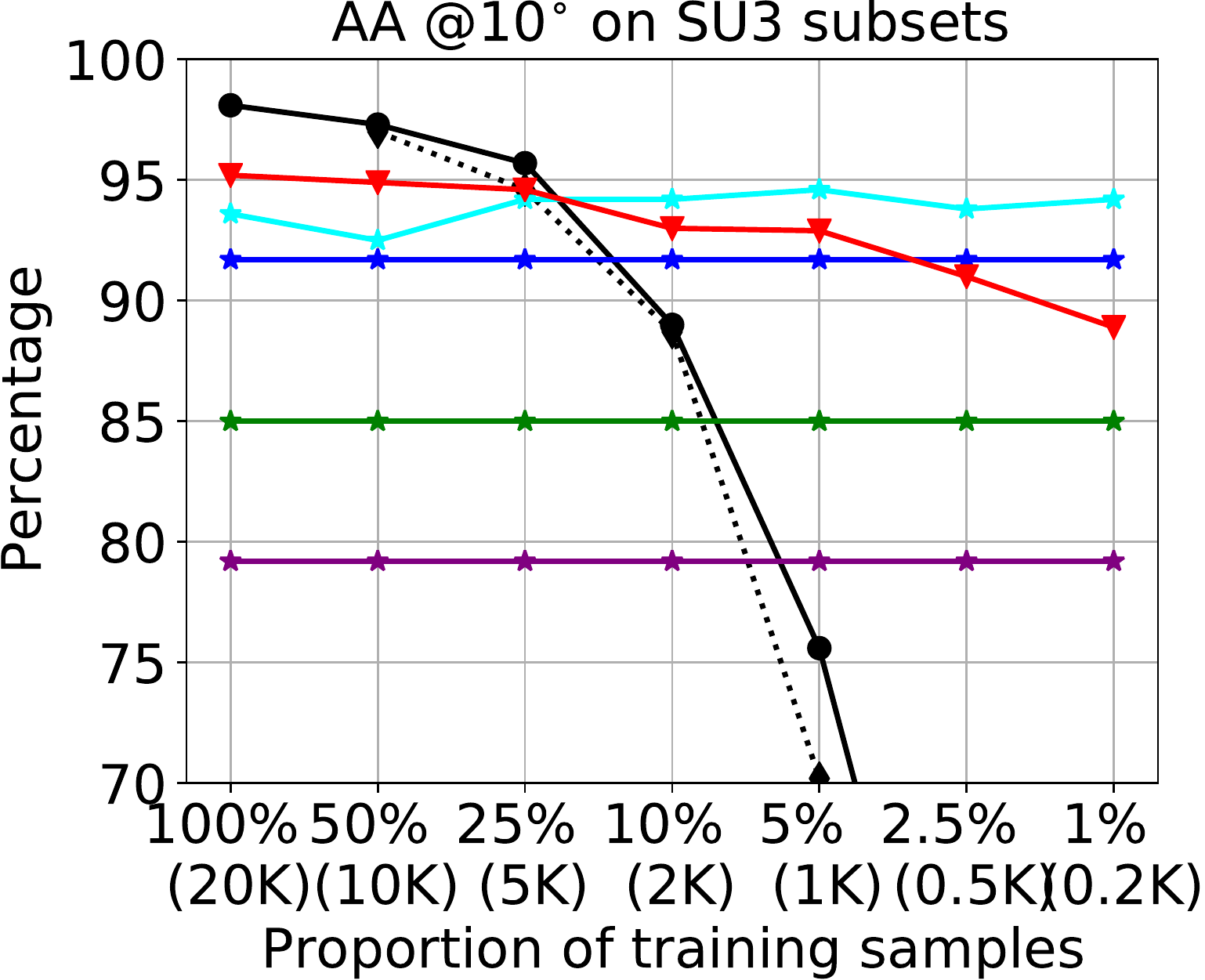} \\
        (a) Data efficiency on ScanNet & (b)  Data efficiency on SU3
    \end{tabular}
    \caption{\small
    \textbf{Exp 3.(a): Reduced data.}
    We report AA @$10^{\circ}$ on various subsets of the ScanNet and SU3 datasets, and indicate the number of parameters in the legend. 
    On the ScanNet dataset, we outperform other methods on the $10\%$, $5\%$ and $2.5\%$ subsets. 
    Our model degrades gracefully when reducing the training samples from 20K to 1K on the SU3 subset, while NeurVPS has a drastic drop in accuracy. 
    CONSAC achieves top results on SU3 due to pre-extracted line segments, but fails on ScanNet because of inaccurate line detection. 
    There is a similar trend for the baselines relying on line segment detection.
    Our model predictions are stable having small variances ($\pm0.50$ and $\pm0.43$ on the $1\%$ subsets of ScanNet and SU3 respectively) across 3 repetitions.
    This experiment validates the data efficiency of our model. 
    }
    \label{fig:exp_subsets}
\end{figure*}
\subsection{Exp 2: Validation on large datasets}
\label{fullset_exps}
We validate that adding prior knowledge does not deteriorate accuracy when there is plenty of data. We compare to five state-of-the-art baselines \cite{feng2010semi, simon2018contrario, li2019quasi, zhou2019neurvps, kluger2020consac} on the ScanNet, SU3 and YUD datasets. 
On the ScanNet and SU3 datasets, we train all learning models from scratch on the full training split. 
On the YUD dataset, we use the pre-trained models on SU3 without fine-tuning. For CONSAC and J-Linkage, we select top-3 predictions. We also measure the inference speed on a single RTX2080 GPU. 
Multi-scale sampling \emph{Ours$^\ast$} achieves 23 FPS, a large speedup over the vanilla design, as we utilize the orthogonality for efficient sampling.

\tab{numerical_eval} shows the AA scores on the ScanNet, SU3 and YUD datasets, while \fig{comparison_all} depicts AA curves for varying angle differences.
The SU3 dataset is easier as most images contain strong geometric cues (e.g. sharp edges and contours); this is no longer the case in the ScanNet dataset. 
The prediction error on the more realistic ScanNet dataset is significantly larger for all methods. 
On the ScanNet dataset, NeurVPS and our model are visibly better than methods relying on predefined line segments as inputs. 
The main advantage of NeurVPS and our model is their ability to learn useful feature representations directly from images. 
On the SU3 dataset, NeurVPS exceeds the other methods in the low-error region (from $0^{\circ}$ to $1^{\circ}$). 
J-Linkage, Quasi-VP and CONSAC have similar results, and all of them stabilize at $1^{\circ}$.
On SU3 our model is less accurate in $0^{\circ}$-$1^{\circ}$, yet it compensates at $\ge 1^{\circ}$. 
Our inferior performance in $0^{\circ}$-$1^{\circ}$ range results from the quantization errors in Hough Transform and the Gaussian sphere mapping. 
On the small-scale YUD dataset \cite{denis2008efficient}, our model achieves comparable accuracy without fine-tuning, and exceeds the other methods in the $\ge 2^{\circ}$ area, indicating the generalization ability of our model in the small data regime. We conclude that our model using prior knowledge performs similar to existing solutions.

\subsection{Exp 3: Challenging scenarios}
\subsubsection{Exp 3.(a): Reduced data}
We evaluate data efficiency by reducing the number of training samples to $\{10\%, 5\%, 2.5\%, 1\% \}$ on the ScanNet dataset, resulting in approximately 20K, 10K, 5K and 2K training images. 
Similarly, we also sample the SU3 dataset into $\{50\%, 25\%, 10\%, 5\%, 2.5\%, 1\% \}$ subsets. We train all learning models from scratch using the default hyperparameters on each subset.

In \fig{exp_subsets} we compare the AA scores at $10^{\circ}$ with state-of-the-art methods. We use our vanilla design without the multi-scale sampling speedup. The first thing to notice is that non-learning methods are robust to data reduction. Yet, non-learning methods cannot take any training data into account, and thus they do not perform as well when more data is available, as we validated in the previous experiment.    
On the ScanNet dataset, our model visibly exceeds the other methods on the $10\%$, $5\%$ and $2.5\%$ subsets. 
In comparison, NeurVPS suffers from large accuracy decreases on small training data subsets. 
When decreasing the number of samples to 2K ($1\%$ subset), we still achieve competitive accuracy when compared to the non-learning methods, while NeurVPS fails to make reasonable predictions due to the lack of data.
This shows the capability of our model to learn from limited data, thanks to the added geometric priors.  

The NeurVPS model has $\times3$ more parameters than our model due to its fully-connected layer with 16M parameters.
For fairness, we also consider `NeurVPS-7M' with reduced fully-connected layers, having a similar number of parameters with our model. 
Both NeurVPS variants perform similar on various subsets.
On the SU3 dataset the accuracy of NeurVPS decreases significantly when reducing the training dataset size, despite its superiority on the large training subsets. 
In comparison, our model degrades gracefully when training data decreases from 20K to 1K. 
Notably, on the $1\%$ subset, with only 200 images for training, we are still able to achieve comparable performance with non-learning methods. 


\begin{figure}[t!]
    \centering
    \begin{tabular}{c}
        \includegraphics[width=0.40\textwidth]{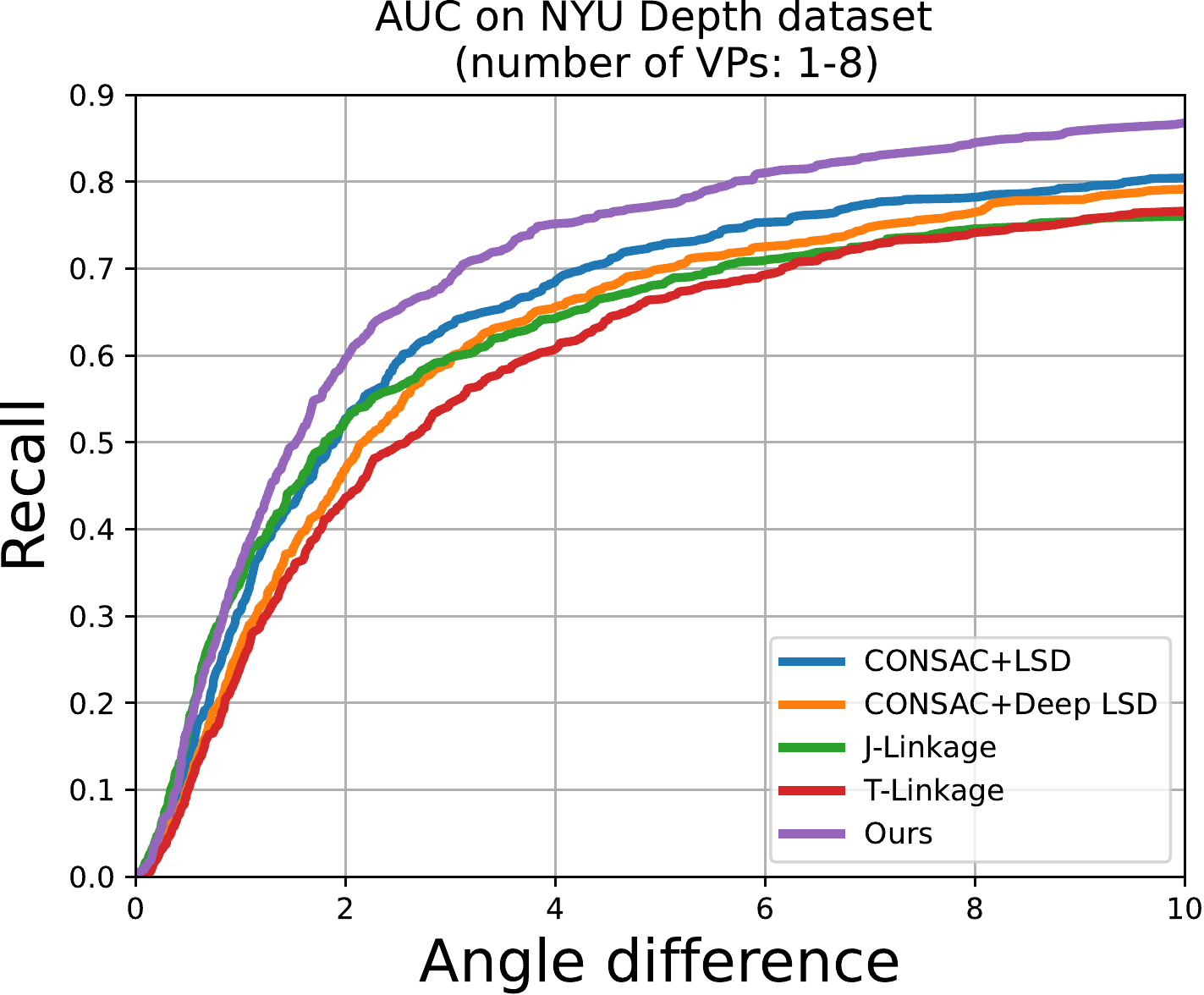} \\
    \end{tabular}
    \caption{\small
    \textbf{Exp 3.(b): Non-Manhattan scenario.}
    We plot the recall curve on the non-Manhattan NYU dataset. 
    Our method outperforms state-of-the-art, illustrating the ability of the model to detect a varying number of vanishing points. See supplementary material for qualitative visualizations.
    }
    \label{fig:comparison_nyu}
\end{figure}
\begin{table}[t!]
    \centering
        \begin{tabular}{p{7.0em}c@{\hskip 0.2in}c@{\hskip 0.2in}c@{\hskip 0.2in}c@{\hskip 0.2in}}
        \toprule
        Datasets &\multicolumn{4}{c}{NYU Depth \cite{nathan2012indoor}}\\ 
        \midrule 
        & \multicolumn{2}{c}{top-$k$ = \#gt} & \multicolumn{2}{c}{top-$k$ = \#pred} \\
       \cmidrule(l){2-3}  \cmidrule(l){4-5}
        AUC & $@5^{\circ}$ & $@10^{\circ}$ & $@5^{\circ}$ & $@10^{\circ}$\\  \midrule
        J-Linkage\cite{feng2010semi}  & 49.30 &  61.28 & 54.48 & 68.34\\
        T-Linkage\cite{magri2014t}  & 43.38 &  58.05 & 47.48 & 64.59\\
        CONSAC\cite{kluger2020consac} &  49.46  & 65.00 & 54.37 & 69.89 \\
        CONSAC\cite{kluger2020consac}+ &  \multirow{2}{*}{46.78}  & \multirow{2}{*}{61.06} & \multirow{2}{*}{49.94} & \multirow{2}{*}{65.96} \\
        DLSD\cite{lin2020deep} &  &  & & \\ 
        VaPiD \cite{liu2021vapid}&  - & 69.10 & - & - \\
        \emph{Ours} &\textbf{55.92}  & \textbf{69.57} & \textbf{57.19}  & \textbf{71.62}\\
        \bottomrule
        \end{tabular} 
    \caption{\small
    \textbf{Exp 3(b): Non-Manhattan scenario.} 
    We report AUC scores on the NYU Depth dataset. 
    Here ``top-$k$ = \#gt" indicates the $k$ most confident predictions where $k$ is the number of annotated instances \cite{kluger2020consac}, while for ``top-$k$ =\#pred" all predictions are used for evaluation.
    Our model exceeds state-of-the-art when detecting a varying number of vanishing points.
    }
    \label{tab:numerical_nyu}
\end{table}

\subsubsection{Exp 3.(b): Non-Manhattan scenario}
We compare with state-of-art methods in a more realistic  non-Manhattan scenario with limited annotated data. 
CONSAC \cite{kluger2020consac} uses the line segment detection in \cite{von2008lsd}. 
We also consider a variant of CONSAC with the more recent line segment detector in \cite{lin2020deep}. 

\fig{comparison_nyu} and \tab{numerical_nyu} display the recall curve and the AUC values on the NYU Depth dataset, respectively. 
Our model consistently outperforms state-of-the-art baselines, and the improvement is more pronounced for larger angular differences. 
Although achieving the second-best result, VaPiD \cite{liu2021vapid} assumes a constant number of instances and requires non-maximum suppression, which often results in over- and under-prediction.
Our model outperforms existing methods by exploiting geometric priors, while not limiting the number of vanishing points detected.

\subsection{Exp 3.(c): Cross-dataset domain switch}

\begin{table}[t!]
    \centering
\resizebox{1\linewidth}{!}{
 \begin{tabular}{p{3.0em}ccccccc}
    \toprule
    \multicolumn{7}{c}{Synthetic - real-world data} \\ \midrule
    Train &  \multicolumn{6}{c}{SU3 \cite{zhou2019learning}}  \\
    \midrule
    Test &  \multicolumn{3}{c}{ScanNet \cite{dai2017scannet}} &  \multicolumn{3}{c}{YUD  \cite{denis2008efficient}} \\
    \cmidrule(l){1-1}\cmidrule(l){2-4}\cmidrule(l){5-7}
    Models &  \emph{Ours} & \footnotesize{NeurVPS} &  \footnotesize{CONSAC} & \emph{Ours} & \footnotesize{NeurVPS} &  \footnotesize{CONSAC} \\
    \midrule
    AA$@3^{\circ}$ & \textbf{15.2} & 11.1 & 10.1& 60.7 & 53.8 & \textbf{61.7} \\  
    AA$@5^{\circ}$ & \textbf{25.9} & 20.3 &17.3& \textbf{74.3} & 65.6 & 73.6 \\  
    AA$@10^{\circ}$ &\textbf{39.5} & 35.5 &27.2 & \textbf{86.3} & 79.7 & 84.4  \\   \bottomrule
    \end{tabular} 
}
\resizebox{1.0\linewidth}{!}{
    \begin{tabular}{p{3.0em}ccccccc}
    \multicolumn{7}{c}{Real-world data} \\ \midrule
    Train &  \multicolumn{6}{c}{NYU Depth \cite{nathan2012indoor}} \\
    \midrule

    Test &  \multicolumn{2}{c}{\footnotesize{ScanNet \cite{dai2017scannet} (AA)}} &  \multicolumn{2}{c}{\footnotesize{YUD \cite{denis2008efficient} (AA)}} &  \multicolumn{2}{c}{\footnotesize{YUD+ \cite{denis2008efficient} (AUC)}}\\
    \cmidrule(l){1-1}\cmidrule(l){2-3}\cmidrule(l){4-5}\cmidrule(l){6-7}
    Models &  \emph{Ours} & \footnotesize{CONSAC} &  \emph{Ours} & \footnotesize{CONSAC} & \emph{Ours} & \footnotesize{CONSAC}  \\
    \midrule

    $@10^{\circ}$ & \textbf{33.6} & 30.3 & \textbf{83.2} & 82.7 & 71.4 & \textbf{75.0}  \\ 
    \bottomrule
    \end{tabular} 
    }
    \caption{\small
    \textbf{Exp 3.(c) Cross-dataset domain switch.} 
    ``Train" and ``Test" specify the training and test datasets. 
    CONSAC uses pre-extracted lines, thus being accurate on YUD/YUD+. 
    However, its accuracy is lower on ScanNet due to the lack of reliable lines. 
    In comparison, \emph{Ours} is more accurate on both ScanNet and YUD without tuning.  
    Our geometric priors improve the transferability of the model across datasets. 
    }
    \label{tab:cross_datasets_yud}
\end{table}

We conduct cross-dataset test on multiple datasets, as displayed in \tab{cross_datasets_yud}. 
We compare with NeurVPS and CONSAC, which achieve top accuracy on individual datasets. When generalizing from synthetic dataset to the real-world (e.g. from SU3 to YUD), our model shows comparative results to CONSAC, which relies on prior line segment detection, making it robust to domain shifts.
We observe a similar trend on real-world datasets (e.g. from NYU to YUD). 
However, on the challenging ScanNet dataset, \emph{Ours} exceeds CONSAC, indicating the advantage of learning semantics over using pre-extracted lines. In contrast, NeurVPS does not transfer well to another dataset. 
This validates the robustness of the two priors in tackling domain shifts.


\section{Conclusion and limitations}
This paper focuses on vanishing point detection relying on well-founded geometric priors.
We add two geometric priors as building blocks in deep neural networks for vanishing point detection: Hough Transform and Gaussian sphere mapping. We validate experimentally the added value of our geometric priors when compared to state-of-the-art Manhattan methods, and show their usefulness on realistic\slash challenging scenarios: with reduced samples, in the non-Manhattan world where the challenge is to predict a varying number of vanishing points without the orthogonality assumption, and across datasets.

\smallskip\noindent\textbf{Limitations.} 
Despite of these improvements, our model also has several limitations.
We pre-compute offline the mapping from images to the Hough bins and to the Gaussian sphere by fixing the size of the Hough histogram, as well as the Fibonacci sampling. 
However, these samplings introduce quantization errors which set an upper bound on accuracy. 
This is the primary reason for the limited accuracy on the SU3 dataset in the low-error region. A future research avenue is exploring an analytical mapping from image pixels to the Gaussian sphere. 
In addition, our model still relies on hundreds of fully labeled samples for training. 
One might consider testing the added geometric priors in an unsupervised or weakly-supervised setting.

\newpage

{\small
\bibliographystyle{ieee_fullname}
\bibliography{egbib}
}
\newpage

\appendix

\section*{\Large\textbf{Supplementary material}}

\section{Multi-scale sampling on the sphere}

Inspired by \cite{zhou2019neurvps}, we use a multi-scale sampling strategy to detect three orthogonal vanishing points in the Manhattan world. We start by uniformly sampling $N_{s=0}$ points at scale $s=0$ on the entire hemisphere.
We input these points into a spherical convolution network. 
Sequentially, we use the Manhattan assumption to choose 3 orthogonal vanishing points with the highest confidence as anchors.
We uniformly sample $N_{s=1}$ points around each anchor in a local neighborhood defined by the radius $\delta_{s=1}$ at scale $s=1$. 
Then, we feed these newly sampled points into a spherical convolution network. 
Finally, the point with highest confidence in each local neighborhood is considered as the anchor for sampling at the $(s+1)$th scale. Specifically, we set $\delta\approx\{90^{\circ}, 13^{\circ}, 4^{\circ}\}$ and $N=\{512, 128, 128\}$. The spherical convolution networks share the same architecture while processing different number of samples. During training, we assign the nearest neighbors to the ground truth as positive samples while the others are considered as negative samples. We compute the cross-entropy losses averaged over positives and negatives respectively, at each scale. 


\section{Datasets}
\begin{table*}[t!]
    \centering
        \begin{tabular}{p{8em} c@{\hskip 0.1in}ccccccc }
        \toprule
        Datasets & Images & Manhattan & Resolution & number of VPs & Training & Validation & Testing\\  \midrule
        SU3   \cite{zhou2019learning} & Synthetic & $\checkmark$ & 512 $\times$ 512 & 3 & $18\,400$ & $2\,300$ & $2\,300$\\
        ScanNet \cite{dai2017scannet} & Real-world & $\checkmark$ & 512 $\times$ 512 &3 &  $189\,916$ & $500$ & $20\,942$ \\
        YUD \cite{denis2008efficient} & Real-world & $\checkmark$ & 480 $\times$ 640 & 3 & 25 & - & 77 \\ 
        \midrule
        NYU Depth \cite{nathan2012indoor} & Real-world & $\times$ & 480 $\times$ 640 & 1-8 & 1000 & 224 & 225 \\ 

        \bottomrule
        \end{tabular} 
        
    \caption{\textbf{Comparison of the four datasets.} The SU3, ScanNet and YUD datasets follow the Manhattan assumption with 3 orthogonal vanishing points, while the NYU Depth dataset is annotated with a varying number of instances. In addition, the size of the four datasets varies substantially. There are only 1000 and 25 training images in NYU and YUD datasets. }
    \label{tab:datasets}
\end{table*}
 

\begin{figure*}[t!]
    \centering
    \includegraphics[width=1.0\textwidth]{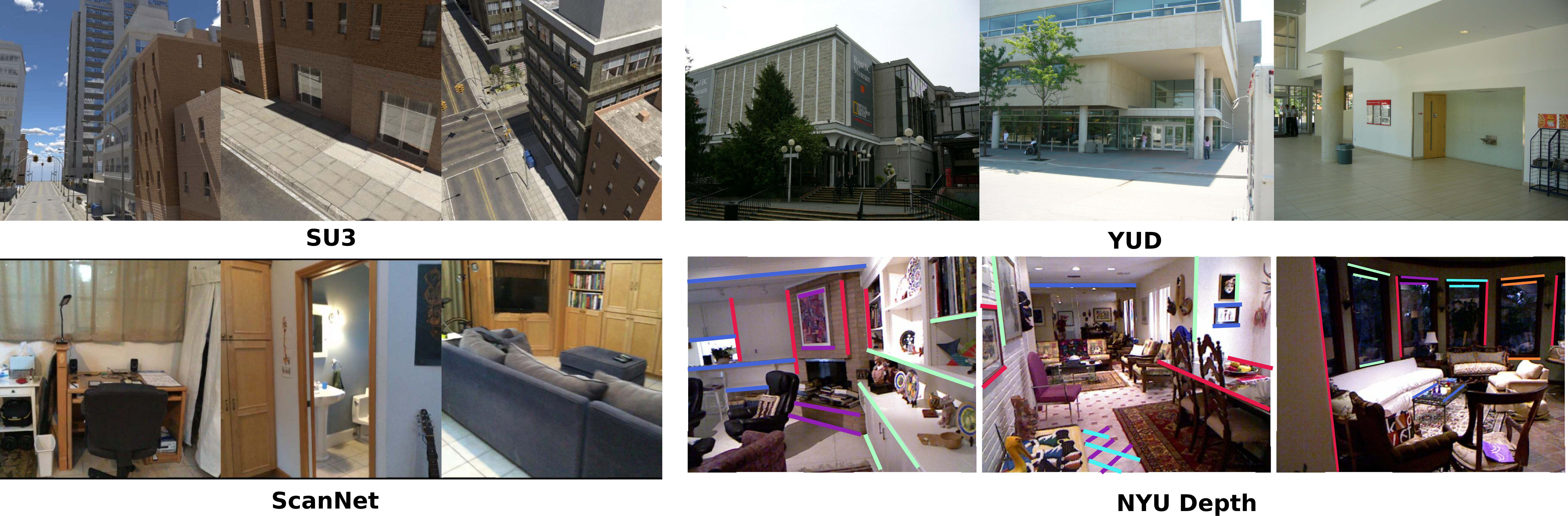}
    \caption{\textbf{Examples from the SU3, ScanNet, YUD and NYU Depth (labeled with ground truth lines) datasets.} 
Images from the SU3 dataset are well-calibrated with clear geometric cues, such as sharp edges and contours. In contrast, the other datasets capture real-world images where image content varies significantly. The NYU Depth dataset is labeled with multiple vanishing points (varying from 1-8). }
    \label{fig:examples}
\end{figure*}

 \tab{datasets} shows a detailed comparison among all datasets, and \fig{examples} displays image examples from each dataset. The SU3 dataset is synthetic and all images are well-calibrated with sharp edges.
 The ScanNet dataset captures indoor scenes in the real-world environments, where image content varies significantly. The YUD dataset captures both indoor and outdoor scenes in urban cities and contains only 102 images. SU3, ScanNet and YUD datasets follow the Manhattan world assumption where there are 3 orthogonal vanishing points. In comparison, the NYU Depth dataset has a varying number of instances across images. Moreover, there are 1449 images in total, and therefore training deep networks is highly challenging on the NYU Depth dataset due to the lack of data.



\section{Visualizations}

\begin{figure*}[t]
    \centering
        \includegraphics[width=0.85\textwidth]{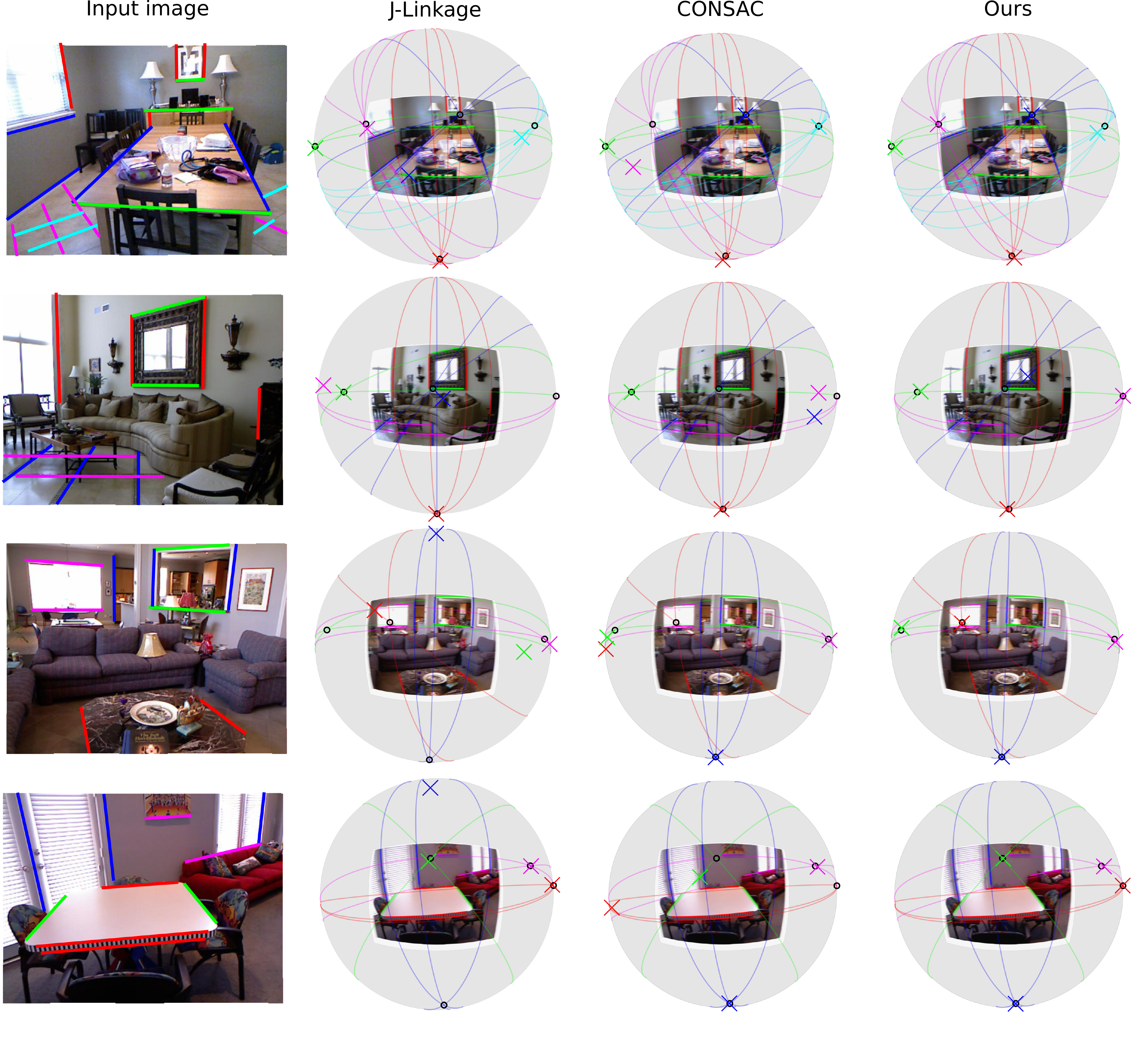}\\
    \caption{\textbf{Visualizations on the NYU Depth dataset.}
    The black $\circ$ represents the ground truth, while the colored $\times$ indicates predictions. 
    Each color corresponds to a set of lines and their related vanishing point.
    Our model is better at localizing multiple vanishing points in the non-Manhattan world, having predictions (colored cross $\times$) closer to the ground truth (black $\circ$), while the predictions of the other methods scatter away from the ground truth, as shown in the first example. 
    }
    \label{fig:vis_nyu}
\end{figure*}

\begin{figure*}[t]
    \centering
        \includegraphics[width=0.9\textwidth]{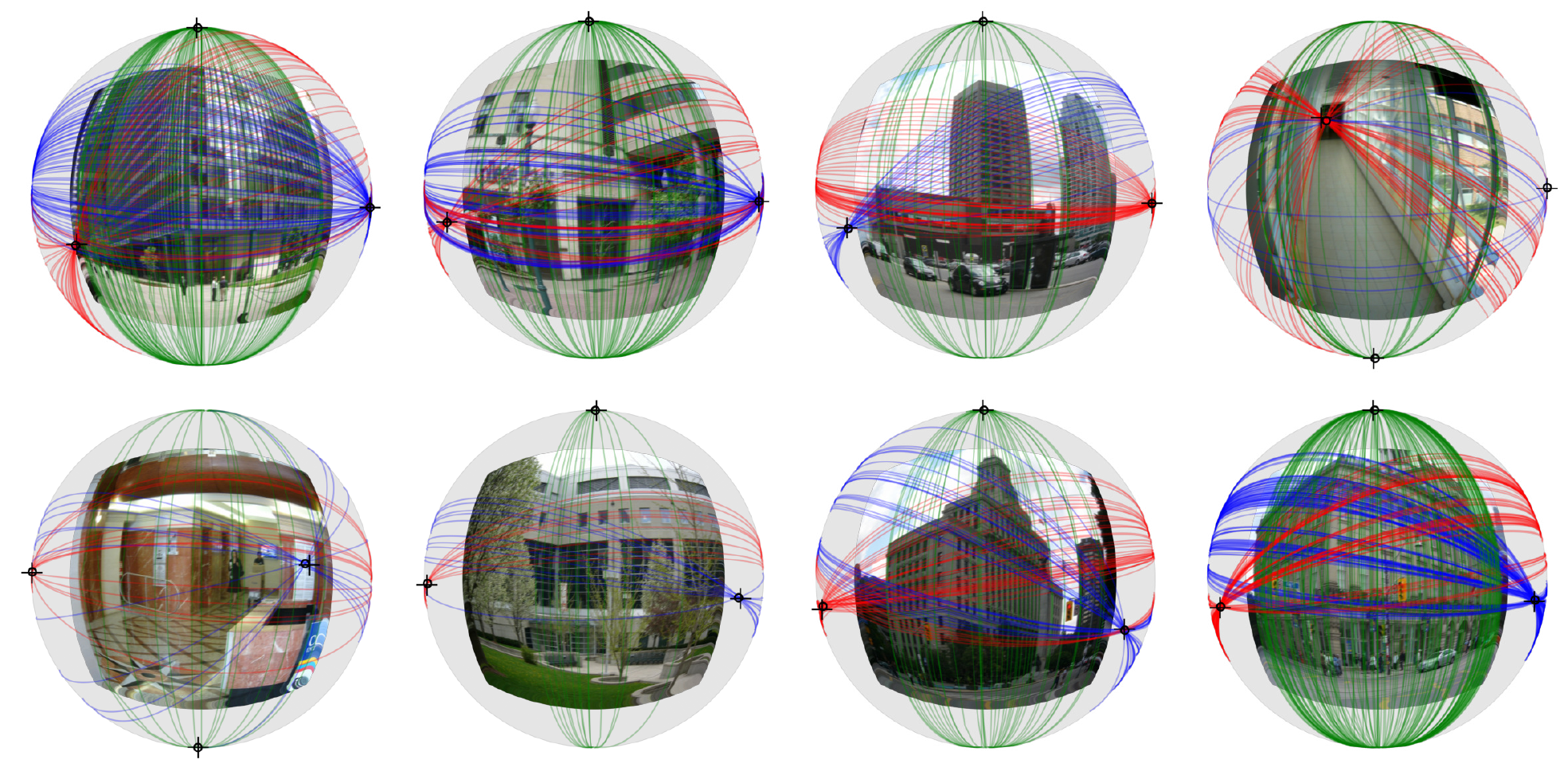}\\
    \caption{\textbf{Visualizations on YUD dataset.}
    We show ground-truth vanishing points ($\circ$) and our predictions ($\times$) on the Gaussian hemisphere, as well as ground truth lines. Our model accurately predicts vanishing points in man-made environments. 
   }
    
    \label{fig:vis_yud}
\end{figure*}

\begin{figure*}[t]
    \centering
        \includegraphics[width=0.9\textwidth]{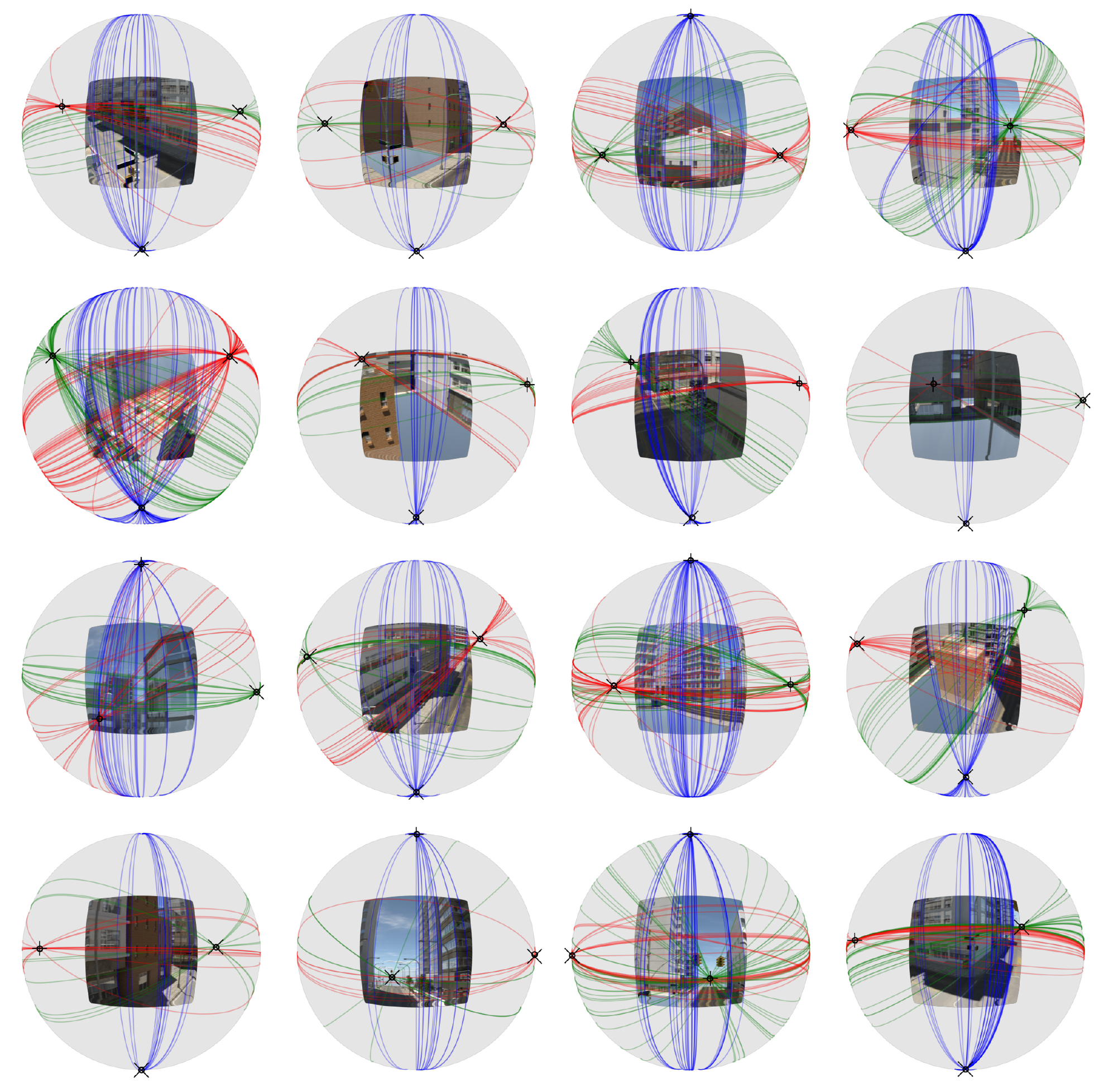}\\
    \caption{\textbf{Visualizations on SU3 dataset.}
    We show ground-truth vanishing points ($\circ$) and our predictions ($\times$) on the Gaussian hemisphere, as well as ground truth lines. Each color represents a cluster of lines that is related to a vanishing point. Our model accurately predicts vanishing points in man-made environments. 
   }
    
    \label{fig:vis_su3}
\end{figure*}

We visualize predictions on the NYU Depth dataset in~\fig{vis_nyu}. 
We show the input images, labeled line segments and detected vanishing points on the hemisphere.
Each color represents a group of lines and their corresponding vanishing point. 
In the third row, our model correctly detects all vanishing points, as the colored $\times$ and $\circ$ overlap. 
In comparison, CONSAC fails to localize the red one and J-Linkage is unable to detect the green one. 
In addition, CONSAC makes nearby predictions: e.g., the blue and pink $\times$ markers in second row. 
This is caused by the LSD \cite{von2008lsd} method producing a large number of outlier segments, resulting in incorrect predictions. 
Our method is suitable for real-world scenarios, where the image content varies substantially.

\fig{vis_yud} and \fig{vis_su3} show detected vanishing points from our model on the SU3 and YUD datasets, respectively. Since all methods make reasonably good prediction and the difference is hardly visible, we only visualize our results.

\begin{figure*}[t]
    \centering
        \includegraphics[width=0.85\textwidth]{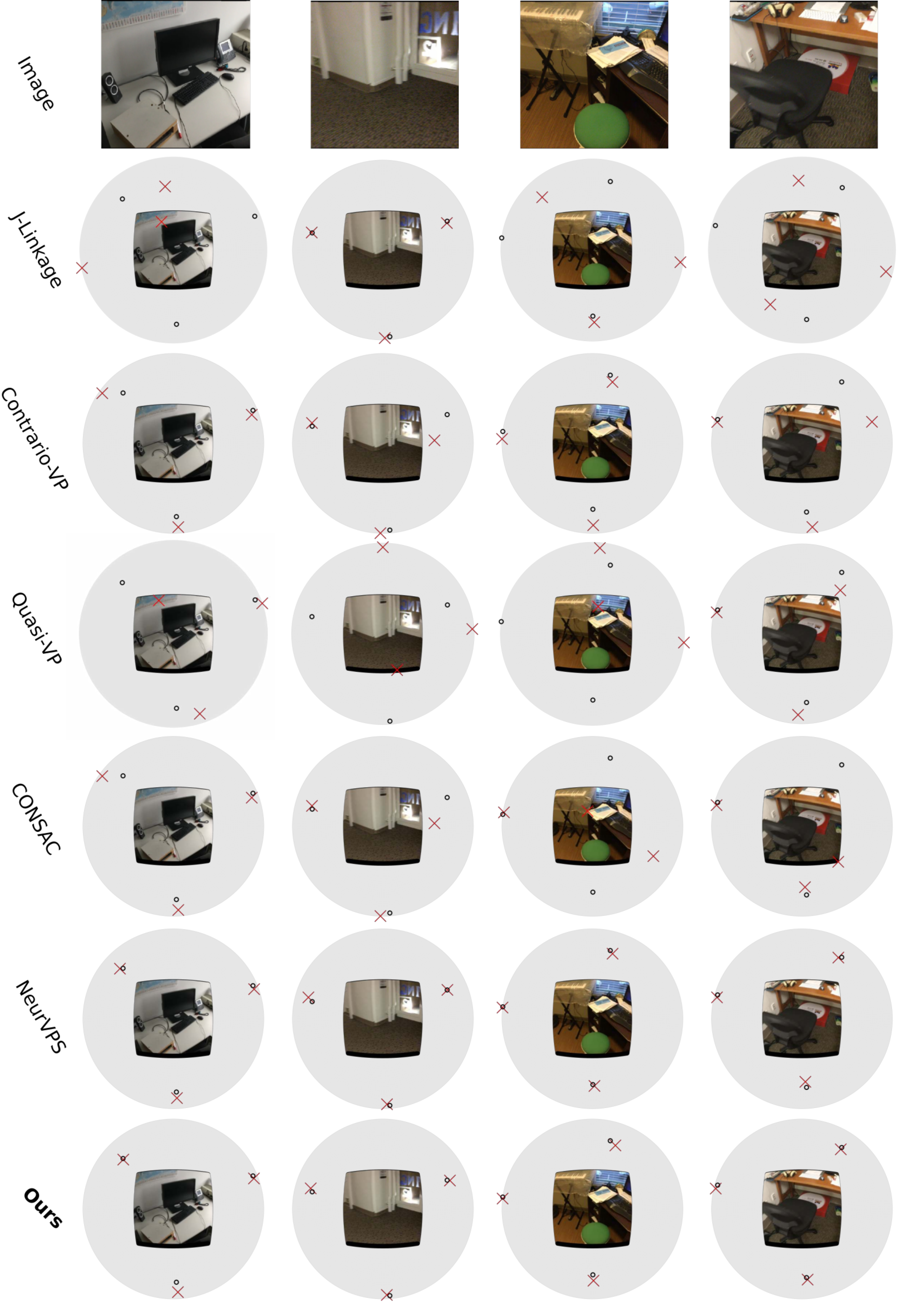}\\
    \caption{\textbf{Visualizations on ScanNet dataset.}
    We show ground-truth vanishing points ($\circ$) and predictions from all baseline methods ($\times$) on the Gaussian hemisphere. Learning-based models shows superior performance to classic line segment-based approaches in complex real-world environments. 
   }
    
    \label{fig:vis_scan}
\end{figure*}

\fig{vis_scan} compares detected vanishing points from all models on the ScanNet dataset. We compare all models in a column-wise manner, where the input image is on the top, while predictions from each method is displayed sequentially. We show the top 3 vanishing points for J-Linkage \cite{feng2010semi} and CONSAC \cite{kluger2020consac}.  In general, NeurVPS and ours are able to localize vanishing points more precisely than other non-learning approaches. As shown in the fourth example where the object is not orthogonally placed, Quasi-VP \cite{li2019quasi} fails due to the presence of strong  outliers and the lack of inliers. This shows the disadvantage of non-learning method in dealing with complicated real-world scenarios. J-Linkage and CONSAC sometimes predict vanishing points far away from the ground truth (e.g., the fourth example), because they are originally designed for multiple vanishing point detection in non-Manhattan world, and do not enforce orthogonality explicitly. Ours show better performance in detecting orthogonal vanishing points from complex scenes thanks to the ability to learn semantic features from images directly in an end-to-end manner.

\end{document}